\documentclass{article}

\usepackage{microtype}
\usepackage{graphicx}
\usepackage{booktabs} 

\usepackage{hyperref}

\usepackage[accepted]{icml2024}

\usepackage{amsmath}
\usepackage{amssymb}
\usepackage{mathtools}
\usepackage{amsthm}
\usepackage{comment}
\usepackage{hyperref}       %
\usepackage{url}            %
\usepackage{booktabs}       %
\usepackage{amsfonts}       %
\usepackage{nicefrac}       %
\usepackage{microtype}      %
\usepackage{color}
\usepackage{amsmath}
\usepackage{amsthm}
\usepackage{graphicx}
\usepackage{tcolorbox}
\usepackage{comment}
\usepackage{tikz}
\usepackage{natbib}
\usepackage{caption}
\usepackage{multirow}
\usepackage{wrapfig}
\usepackage{amsfonts, amsmath}
\usepackage{enumitem}

\usepackage{tabularx}

\usepackage{varwidth}
\usepackage{lipsum}
\usepackage{multicol}
\newsavebox{\algbox}
\usepackage{import}
\usetikzlibrary{bayesnet}
\usetikzlibrary{arrows}
\usepackage{makecell}

\usepackage[utf8]{inputenc} 
\usepackage[T1]{fontenc}    
\usepackage{hyperref}       
\usepackage{url}            
\usepackage{booktabs}       
\usepackage{amsfonts}       
\usepackage{nicefrac}       
\usepackage{microtype}      
\usepackage{url}            %
\usepackage{booktabs}       %
\usepackage{amsfonts}       %
\usepackage{nicefrac}       %
\usepackage{microtype}      %
\usepackage{color}
\usepackage{amsmath}
\usepackage{amsthm}
\usepackage{graphicx}
\usepackage{comment}
\usepackage{tikz}
\usepackage{natbib}
\usepackage{caption}
\usepackage{multirow}
\usepackage{wrapfig}
\usepackage{amsfonts, amsmath}
\usepackage{enumitem}
\usepackage{soul}
\usepackage{bm}
\usepackage{wrapfig}
\usepackage{listings}
\usepackage[title]{appendix}
\usepackage[bottom]{footmisc}
\usepackage[makeroom]{cancel}
\usepackage[normalem]{ulem}
\setcitestyle{authoryear,open={(},close={)}}
\usepackage{array}
\newcolumntype{P}[1]{>{\centering\arraybackslash}p{#1}}
\usepackage{varwidth}
\usepackage{lipsum}
\usepackage{multicol}
\usepackage{import}
\usetikzlibrary{bayesnet}
\usetikzlibrary{arrows}
\usepackage{booktabs}

\usepackage[utf8]{inputenc} 
\usepackage[T1]{fontenc}    
\usepackage{hyperref}       
\usepackage{url}            
\usepackage{booktabs}       
\usepackage{amsfonts}       
\usepackage{nicefrac}       
\usepackage{microtype}      
\usepackage{xcolor}         

\def\x{\mathbf{x}}
\def\e{\mathbf{e}}
\def\bfx{\mathbf{X}}

\def\r{\mathbf{r}}

\newcommand{\yv}[1]{\textcolor{olive}{[YV: #1]}}
\usepackage[capitalize,noabbrev]{cleveref}
\usepackage{amsmath}
\usepackage{empheq}
\newtcbox{\mymath}[1][]{%
    nobeforeafter, tcbox raise base, colframe=teal!60!black,
    colback=teal!10, boxrule=1pt,
    #1}

\definecolor{lb}{RGB}{31,119,180}

\usepackage{tcolorbox}
\newtcolorbox{mybox}[1]{colback=lb!5!white,colframe=lb!70!black,fonttitle=\bfseries,title=#1}
\usepackage{pifont}
\newcommand{\cmark}{\textcolor{green!60!black}{\ding{51}}}
\newcommand{\xmark}{\textcolor{red!60!black}{\ding{55}}}

\usepackage{colortbl}
\tcbuselibrary{skins}
\tcbset{tab2/.style={enhanced,fonttitle=\bfseries,
colback=white,colframe=gray,colbacktitle=white,
coltitle=black,center title}}

\newtheorem{definition}{Definition}
\newtheorem{proposition}{Proposition}

\newcommand{\multiset}[1]{
  \{\!\!\{#1\}\!\!\}
}

\usepackage[textsize=tiny]{todonotes}

\icmltitlerunning{Topological Neural Networks go Persistent, Equivariant, and Continuous}

\begin{document}

\twocolumn[
\icmltitle{Topological Neural Networks go Persistent, Equivariant, and Continuous}




\icmlsetsymbol{equal}{*}

\begin{icmlauthorlist}
\icmlauthor{Yogesh Verma}{yyy}
\icmlauthor{Amauri H. Souza}{yyy,abc}
\icmlauthor{Vikas Garg}{yyy,comp}
\end{icmlauthorlist}

\icmlaffiliation{yyy}{Department of Computer Science, Aalto University, Finland}
\icmlaffiliation{abc}{Federal Institute of Ceará}
\icmlaffiliation{comp}{YaiYai Ltd}

\icmlcorrespondingauthor{Yogesh Verma}{yogesh.verma@aalto.fi}

\icmlkeywords{Machine Learning, ICML}

\vskip 0.3in
]




\printAffiliationsAndNotice{} 
\begin{abstract}

Topological Neural Networks (TNNs) incorporate higher-order relational information beyond pairwise interactions, enabling richer representations than Graph Neural Networks (GNNs). Concurrently, topological descriptors based on persistent homology (PH) are being increasingly employed to augment the GNNs. We investigate the benefits of integrating these two paradigms.     
Specifically, we introduce {\em TopNets} as a broad framework that subsumes and unifies various methods in the intersection of GNNs/TNNs and PH such as (generalizations of) RePHINE and TOGL. TopNets can also be readily adapted to handle (symmetries in) geometric complexes, extending the scope of TNNs and PH to spatial settings. Theoretically, we show that PH descriptors can provably enhance the expressivity of simplicial message-passing networks. Empirically, (continuous and $E(n)$-equivariant extensions of) TopNets achieve strong performance across diverse tasks, including antibody design, molecular dynamics simulation, and drug property prediction.   

\end{abstract}

\begin{table*}[!t]
\centering
\begin{minipage}{\textwidth}
\centering
\small
    \caption{\textbf{Overview of recent methods for relational data and summary of our contributions}. E: Equivariant, P: Persistent, C: Continuous, and HO: higher order.}
\begin{tcolorbox}[tab2,tabularx={llccccccll}, boxrule=1pt,top=0.9ex,bottom=0.9ex,colbacktitle=lb!15!white,colframe=lb!70!white]

  & \multicolumn{5}{c}{
    {{\bf Recent methods for relational data }}
  }
  &
  & \multicolumn{3}{c}{
    {\multirow{2}{*}{\bf Main contributions of this work}} 
  }

  \\
  \cmidrule{2-6} 
               &  Method & E & P & C & HO  \\ \cmidrule{2-6} \cmidrule{9-10}
               & TOGL~\citep{horn2021topological}      & \xmark & \cmark & \xmark& \xmark   & & & \textbf{Section 3}  \\
             &   PersLay~\citep{carrierePersLayNeuralNetwork2020}     & \xmark & \cmark & \xmark& \xmark   & && \quad Unified Framework: TopNets  \\
              &   RePHINE~\citep{rephine}     & \xmark & \cmark & \xmark& \xmark  & & & \quad TNNs + PH $\succ$ TNNs & Prop. 1  \\
             &    MPSN~\citep{bodnar2021weisfeiler}  & \xmark & \xmark & \xmark & \cmark & & & \textbf{Section 4}  \\ 
              &   CWN~\citep{Bodnar2021}  & \xmark & \xmark & \xmark & \cmark & & & \quad $E(n)$-Equivariant TopNets (E-TopNets) \\
              &   CAN~\citep{giusti2023cell}  & \xmark & \xmark & \xmark & \cmark &&  &\quad Invariant persistence diagrams & Prop. 2  \\
              &   IMPSN~\citep{eijkelboom2023mathrm}     & \cmark & \xmark & \xmark& \cmark   &&&\textbf{Section 5}    \\
              &   EGNN~\citep{satorras2021n}     & \cmark & \xmark & \xmark& \xmark    & & & \quad Continuous (Equivariant) TopNets \\
              &   E3NN~\citep{geiger2022e3nn}     & \cmark & \xmark & \xmark& \xmark     &&& \quad Discretization error (TOGL) & Prop. 3  \\
              &   GATr~\citep{brehmer2023geometric} & \cmark & \xmark & \xmark& \xmark  &&& \quad Discretization error (RePHINE) & Prop. 4 \\
             &    GRAND~\citep{chamberlain2021grand}& \xmark & \xmark & \cmark & \xmark  &&& \textbf{Section 6} \\
            &     GREAD~\citep{choi2022gread}& \xmark & \xmark & \cmark & \xmark & && \quad  \multirow{2}{*}{\makecell{Experiments: graph classification, drug \\ property prediction, and generative design}}\\
            &     GRAND++~\citep{thorpe2022grand++} & \xmark & \xmark & \cmark & \xmark & &&   \\
                \cmidrule{2-6}
             &    \textbf{TopNets (ours)}    & \cmark & \cmark & \cmark &\cmark  && &  \\
\end{tcolorbox}

    \label{tab:overview}
\end{minipage}
\end{table*}

\section{Introduction}
Relational data in diverse settings such as social networks \citep{freeman2004development}, and proteins \citep{jha2022prediction} can be effectively abstracted via graphs. GNNs have enabled considerable success in representing such data \citep{bronstein2021geometric}. However, their limitations such as inability to distinguish non-isomorphic graphs and compute graph properties \citep{gin,weisfeiler1968reduction, gnns_repr} have spurred research efforts toward designing more powerful models that can leverage higher-order interactions, e.g., hierarchical \textit{part-whole} relations.

Topological deep learning (TDL) \citep{Papillon23} views graphs as 1-dimensional simplicial complexes, and employs general abstractions to process data with higher-order relational structures. TNNs, a broad class of topological neural architectures, have yielded state-of-the-art performance on various machine learning tasks \citep{dong2020hnhn,chen2019mixed,barbarossa2020topological}, showcasing high potential for numerous applications.     



Simultaneously, descriptors based on PH \citep{horn2021topological,carrierePersLayNeuralNetwork2020,rephine}, a workhorse from topological data analysis (TDA), capture important topological information such as the number of components and independent loops. Augmenting GNNs with persistent features affords powerful representations. However, the merits of integrating persistence in TNNs remain unexplored. In particular, numerous real-world tasks involving topological objects exhibit symmetries under the Euclidean group $\mathrm{E}(n)$, such as translations, rotations, and reflections. Examples range from predicting molecular properties \citep{ramakrishnan2014quantum}, 3D atomic systems \citep{duval2023hitchhikers}, to generative design and beyond. While various approaches use these symmetries effectively, including Tensor Field Networks \citep{thomas2018tensor}, $\mathrm{SE}(3)$ Transformers \citep{fuchs2020se}, EGNNs \citep{satorras2021n}, and EMPSNs \citep{eijkelboom2023mathrm}, their expressivity remains limited as they fail to capture certain topological structures \citep{joshi2023expressive} in geometrical simplicial complexes.

We strive to bridge this gap with a general recipe to leverage the best of both worlds.  Specifically, we propose TopNets (\textbf{To}pological \textbf{P}ersistent Neural \textbf{Net}work\textbf{s}) as a comprehensive framework unifying 
TNNs and PH. Our approach allows us to seamlessly accommodate additional contextual cues; e.g., TopNets can process spatial information via geometric color filtrations.  We analyze TopNets from both theoretical and practical perspectives, illuminating their promise across diverse tasks. 

We reinforce the versatility of TopNets by designing their continuous counterparts, defining associated Neural ODEs over simplicial complexes and elucidating error bounds between the discrete and continuous systems. We thus build on the remarkable success of Neural ODEs \citep{ODE, kim2023trainability,marion2023generalization} across various domains, including spatiotemporal forecasting \citep{yildiz2019ode2vae,li2020fourier,lu2021learning,kochkov2021machine,brandstetter2022clifford,verma2024climode}, generative modeling \citep{grathwohl2018ffjord,lipman2023flow,modflow,pmlr-v202-verma23a}, and graph representation learning \citep{poli2019graph,iakovlev2020learning,chamberlain2021grand,thorpe2022grand++,choi2022gread}. 

We summarize our main contributions below:
\begin{enumerate}
    \item (\textbf{Methodology}) we propose TopNets, a general unifying framework that combines TNN with PH and leverages persistent homology to boost the expressivity of (equivariant) message-passing simplicial networks;
    \item  (\textbf{Theory}) we derive a set of associated Neural-ODEs for various TNNs and PH over simplicial complexes and compute the associated discretization error bound between discrete and continuous systems;
    \item (\textbf{Empirical}) TopNets achieve strong performance across diverse real-world tasks such as graph classification, drug property prediction, and generative design.\footnote{Code is available here: \url{https://github.com/Aalto-QuML/TopNets} }
\end{enumerate}

We compare TopNets with several other recent methods for modeling relational data in \autoref{tab:overview}.

\color{black}

%
%

\section{Background}
We begin with notions from topological ML, persistent homology, equivariance, and Graph ODEs that we use.


%
%
\begin{figure*}[!tb]
    \centering
    \includegraphics[width=\textwidth]{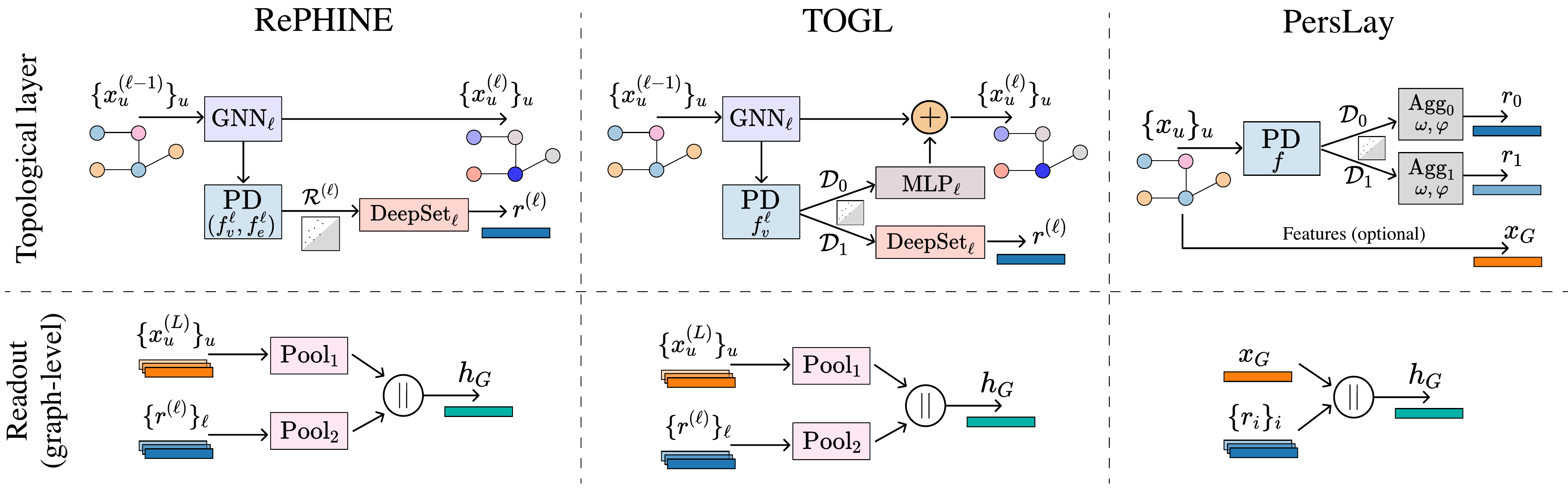}
    \caption{\textbf{Comparison of representative PH-based architectures for graph learning.}}
    \label{fig:model_arch_v2}
\end{figure*}

\textbf{Simplicial complexes.} An  \emph{abstract simplicial complex} (ASC) over a vertex set $V$ is a set $K$ of subsets of $V$ (called \emph{simplices}) such that, for every $\sigma \in K$ and every non-empty $\tau \subset \sigma$, we have that $\tau \in K$.
Let $\sigma$ be a simplex, then its non-empty subsets $\tau \subset \sigma$ are called \emph{faces}, and $\sigma$ is a \emph{coface} of $\tau$. The dimension of a simplex is equal to its cardinality minus $1$, and the dimension of a simplicial complex is the maximal dimension of its simplices.
We denote by $K_{[i]}$ the subset of $i$-dim simplices of $K$.
Here, we represent simplices using square brackets. For instance, $K=\{[0], [1], [0, 1]\}$ denotes a $1$-dim simplicial complex over $V=\{0,1\}$, and the $0$-dim simplices $[0]$ and $[1]$ are the faces of the simplex $[0, 1]$.
%

We also consider simplicial complexes with features. In particular, a \emph{geometric simplicial complex} is a tuple ($K$, $x$, $z$) where $x: K \rightarrow \mathbb{R}^{d_x}$ and $z: K_{[0]} \rightarrow \mathbb{R}^{d_z}$ are functions that assign to a simplex $\sigma$ an attribute (or color) $x(\sigma)$ and a geometric feature $z(\sigma)$, respectively. For convenience, hereafter, we denote the feature vectors of $\sigma$ by $x_\sigma$ and $z_\sigma$.

\paragraph{Graph neural networks (GNNs).} Let $G=(V, E)$ be an undirected graph with vertex set $V$ and edge set $E \subseteq V \times V$ --- note that graphs are 1-dim ASCs. 
To obtain meaningful graph representations, message-passing GNNs~\citep{gilmer2017neural,gin,velivckovic2017graph} employ a sequence of message-passing steps, where each node $v$ aggregates messages from its neighbors $\mathcal{N}(v)=\{u: (v, u) \in E\}$ and use the resulting vector to update its own embedding. In particular, starting from $x_v^{0}=x_v$ $\forall v \in V$, GNNs recursively apply the update rule
\begin{equation*}
x_v^{\ell+1} = \mathrm{Upd}_\ell\left(x_v^{\ell}, \mathrm{Agg}_\ell (\multiset{x^{\ell}_u : u \in \mathcal{N}(v)}) \right),
\end{equation*}
where $\multiset{\cdot}$ denotes a multiset, $\mathrm{Agg}_\ell$ is an order-invariant function and $\mathrm{Upd}_\ell$ is an arbitrary update function.

\paragraph{Topological neural networks} \citep[TNNs, e.g.,][]{Bodnar2021,Hensel21,Hofer2017} consist of neural models for processing data with high-order relational structure. 
\citet{Papillon23} provide a unified framework to describe message-passing TNNs --- here we focus on models for simplicial complexes. After specifying \emph{neighborhood structures}, which define how simplices (possibly of different dimensions) can locally interact, TNNs recursively update the simplices' embeddings via message passing. This general message-passing procedure comprises: $i$) message computation, $ii$) within-neighborhood aggregation, $iii$) between-neighborhood aggregation, and $iv$) update. More specifically, let $\mathcal{N}$ define a neighborhood structure. For each simplex $\sigma \in K^{\ell}$ at layer $\ell$, we compute the messages $
m^{\ell, \mathcal{N}}_{\sigma' \rightarrow \sigma} = \mathrm{Msg}_{\ell, \mathcal{N}}(x^{\ell}_\sigma, x^{\ell}_{\sigma'})
$
from all $\sigma' \in \mathcal{N}(\sigma)$, where $\mathrm{Msg}_{\ell, \mathcal{N}}$ is an arbitrary function. Then, the messages to simplex $\sigma$ are aggregated, that is,
\begin{align}
    m^{\ell, \mathcal{N}}_\sigma &= \mathrm{WithinAgg}_\ell(\{ m^{\ell, \mathcal{N}}_{\sigma' \rightarrow \sigma} : \sigma' \in \mathcal{N}(\sigma)\}), \\
        m^{\ell}_\sigma &= \mathrm{BetweenAgg}_\ell(\{ m^{\ell, \mathcal{N}}_\sigma : \mathcal{N} \in \mathcal{N}_\text{all}\}),
\end{align}
where $\mathcal{N}_\text{all}$ is a set of neighborhoods comprising, e.g., co-boundary, boundary, lower-, and upper- adjacencies \citep{bodnar2021weisfeiler}. 
Finally, we apply a function $\mathrm{Update}_\ell$ to obtain the refined feature vector at layer $\ell+1$ as 
\begin{align}
    x^{\ell+1}_\sigma = \mathrm{Update}_\ell(
    m^{\ell}_\sigma, x^{\ell}_\sigma).
\end{align}
Notably, TNNs subsume a large class of models, including message-passing GNNs.

\paragraph{Persistent homology.} A \emph{filtration} of a simplicial complex $K$ is a finite nested sequence of subcomplexes of $K$, i.e., $\emptyset = K_0 \subset K_1 \subset ... \subset K$.
To obtain a valid filtration, it suffices to ensure that all the faces of a simplex $\sigma$ do not appear later than $\sigma$ in the filtration.
To achieve that, a typical choice consists of defining a filtering (or filtration) function $f$ on the vertices of the simplicial complex, and use it to rank each simplex $\sigma \in K$ as $o_f(\sigma)=\max_{v \in \sigma}f(v)$. Let $\alpha_1 < \dots < \alpha_n$ be an increasing sequence of vertex filtered values, i.e., $\alpha_i \in \{ f(v): v \in K_{[0]}\}$; then, we index the filtration steps using real numbers and define the filtration of $K$ induced by $f$ as $K_{\alpha_i} = \{\sigma \in K: o_f(\sigma) \leq \alpha_i\}$ for $i=1, \dots, n$.
%
%
Another common strategy adopts filtering functions on vertex features $x_v$ and redefine $o_f(\sigma; x)=\max_{v \in \sigma}f(x_{v})$. Filtrations induced by functions on vertex features (or colors) are called \emph{vertex-color filtrations}.

The idea of persistent homology (PH) is to keep track of the appearance and disappearance of topological features (e.g., connected components, loops, voids) in a filtration. 
If a topological feature first appears in $K_{\alpha_i}$ and disappears in $K_{\alpha_j}$, then we encode its persistence as a pair $(\alpha_i, \alpha_j)$; if a feature does not disappear, then its persistence is $(\alpha_i, \infty)$.
The collection of all pairs forms a multiset that we call \emph{persistence diagram}.
We use $\mathcal{D}_{i}$ to denote the persistence diagram for $i$-dim topological features.
We provide more details in \autoref{sec:ph}.

Persistence diagrams are usually vectorized before being combined with ML models. In this regard, \citet{carrierePersLayNeuralNetwork2020} proposed a general framework, called PersLay, that computes a vector representation for a given diagram $\mathcal{D}$ as
$$
\mathrm{Agg}\left(\{\omega(p) \varphi(p) : p \in \mathcal{D}\}\right),
$$
where $\mathrm{Agg}$ is a permutation invariant operation (e.g., mean, maximum, sum), $\omega: \mathbb{R}^2 \mapsto \mathbb{R}$ is an arbitrary function that assigns a weight to each persistence pair, and $\varphi : \mathbb{R}^2 \mapsto \mathbb{R}^q$ maps each pair to a higher dimensional space.
Notably, PersLay introduces choices for $\varphi$ that generalize many vectorization methods in the literature \citep[e.g.,][]{deepsets,bubenikStatisticalTopologicalData2015,adamsPersistenceImagesStable2016,kusanoPersistenceWeightedGaussian2016}.

\paragraph{Combining PH and GNNs.} Recently, PH has been used to boost the expressive power of GNNs. \citet{horn2021topological} introduce TOGL --- a general approach for incorporating topological features from PH into GNN layers. In particular, TOGL leverages node embeddings at each layer of a GNN to obtain vertex-color filtrations. The 0-dim individual persistence tuples are vectorized using MLPs and added to the corresponding node features at each layer. For 1-dim tuples, TOGL applies DeepSets to get a graph-level vector that runs through the final fully-connected layers of the GNN.

\citet{rephine} use independent vertex-color and edge-color filtering functions to obtain more expressive persistent diagrams called RePHINE. More specifically, RePHINE first computes persistence diagrams from a filtration induced by edge colors. Each tuple of the diagram is then augmented based on the vertex colors and the local edge-color information around each vertex. RePHINE diagrams are vectorized using DeepSets and combined with graph-level GNN embeddings in the final classifier. \autoref{fig:model_arch_v2} depicts the architectures of RePHINE, TOGL, and PersLay.


\paragraph{$\mathbf{E(n)}$-Equivariant networks.} 

Let $\mathfrak{G}$ be a group acting on two sets $\mathcal{X}$ and $\mathcal{Y}$. We say a function $f: \mathcal{X} \rightarrow \mathcal{Y}$ is $\mathfrak{G}$-equivariant if it commutes with the group actions, i.e., for all $\mathfrak{g} \in \mathfrak{G}$ and $x \in \mathcal{X}$, we have that  $f(\mathfrak{g} \cdot x)= \mathfrak{g} \cdot f(x)$. 
Here, we are interested in models on geometric simplicial complexes that are equivariant to the Euclidean group $E(n)$, which comprises all translations, rotations, and reflections of the $n$-dim Euclidean space. \citet{eijkelboom2023mathrm} introduce Equivariant Message-Passing Simplicial Networks (EMPSNs), which extends the $E(n)$-equivariant GNNs \citep{satorras2021n} to geometric simplicial complexes. 
For each simplex $\sigma \in K^{\ell}$ at layer $\ell$, EMPSNs compute the messages $m^{\ell, \mathcal{N}}_{\sigma' \rightarrow \sigma} = \mathrm{Msg}_{\ell, \mathcal{N}}(x^{\ell}_\sigma, x^{\ell}_{\sigma'},\mathrm{Inv}(\sigma,\sigma'; z^\ell) )$ from all $\sigma' \in \mathcal{N}(\sigma)$, where $\mathrm{Inv}(\sigma,\sigma'; z^\ell)$ denotes invariant features (e.g., volumes, angles, distances) computed using coordinates from $z^\ell$. Then, the messages to simplex $\sigma$ are aggregated using $\mathrm{WithinAgg}_\ell$ and $\mathrm{BetweenAgg}_\ell$ the same way as in TNNs to obtain an aggregated message $m^\ell_\sigma$. Finally, we recursively update the features and coordinates as 
\begin{align}
    \!\!\!x^{\ell+1}_\sigma &= \mathrm{Update}_\ell(
    m^{\ell}_\sigma, x^{\ell}_\sigma) \\
    \!\!z^{\ell+1}_\sigma &= C \!\! \sum_{\sigma' \in \mathcal{N}_\uparrow(\sigma)}\!\!\! (z^{\ell}_{\sigma} - z^{\ell}_{\sigma'}) \phi_{z}^{\ell}(m^{\ell, \mathcal{N}_\uparrow}_{\sigma' \rightarrow \sigma})~~\forall\sigma \in K_{[0]}
\end{align}
where $\mathcal{N}_\uparrow$ denotes the upper-adjancency, $C$ is a normalization constant, and $\phi^\ell_z$ is an arbitrary function.

\paragraph{Graph ODEs.} 
Neural Ordinary Differential Equations (ODEs) represent a class of implicit deep learning models characterized by an ODE, where the vector field is parameterized by a neural network \citep{ode1,ode2,ODE,ode3}. Graph ODEs~\citep{poli2019graph} generalize Neural ODEs to garphs. For instance, we can track the evolution of signals defined over the vertices of a graph as a differential equation
\begin{align}
    \dot{x}_{v} = \frac{dx_{v}}{dt} = f(t,x_{v}, \{x_u\}_{u \in \mathcal{N}(v)}).
\end{align}
Here, the vector field $f$ is parameterized by a neural network. A notable feature is that, under a mild assumption on $f$, employing an Euler scheme for $N$ time-steps converges to an $N$-layer Graph ResNet \citep{sander2022residual}. This convergence implies that Graph ODEs inherently inherit the capability to incorporate relational inductive biases seen in GNNs  while maintaining the dynamic system perspective of continuous-depth models. The versatility of Graph ODEs has paved the way for the design of novel graph neural networks, such as GRAND \cite{chamberlain2021grand}, GREAD \citep{choi2022gread}, and AbODE \citep{pmlr-v202-verma23a}. 



\section{A unified framework: Topological persistent neural networks (TopNets)}
\label{sec:unify}
We now introduce a general framework that combines TNNs and PH for expressive learning on topological objects. We call this framework \emph{topological persistent neural networks} or \textcolor{black}{TopNets}, in short. Notably, we show that TopNets subsume several methods at the intersection of PH and GNNs.

To motivate our framework, we show that persistent homology features bring in additional expressive power to TNNs. \citet{bodnar2021weisfeiler} introduce a Simplicial Weisfeiler-Leman (SWL) test to characterize the expressivity of simplicial message-passing networks (SMPNs) --- a general TNN for simplicial complexes. They show that SWL (with clique complex lifting) is strictly more powerful than 1-WL. Our next result (\autoref{prop:TNNs_PH}) implies that the combination of SWL and PH is strictly more expressive than the SWL test. 

\begin{proposition}[SWL + PH $\succ$ SWL]\label{prop:TNNs_PH} 
There are pairs of non-isomorphic clique complexes that SWL cannot distinguish but persistence diagrams from color-based filtrations can.
\end{proposition}

Prior works \citep{horn2021topological,Rieck2023,rephine} have demonstrated that PH can be used to increase the power of GNNs. \autoref{prop:TNNs_PH} shows that this also applies to TNNs on simplicial complexes. 

Given an input simplicial complex, each layer in a TopNet first applies a general message-passing (MP) procedure to obtain a refined attributed complex, as in TNNs. Then, we compute persistence diagrams followed by a vectorization scheme that assigns each simplex a topological embedding. Next, TopNets obtain two complex-level representations: the first consists of a joint MP-PH vector derived from a combination of the features of the complex and the topological embeddings; and the second one is obtained by merging the PH-based descriptor associated with each simplex via an order-invariant function. Finally, we combine the representation of the simplices at each layer and dimension, and  apply two readout layers. The first aims to combine information from different layers (but same dimension) while the second readout function further processes the resulting representations across dimensions. In the following, we formalize these steps.

\begin{mybox}{Steps of a TopNet layer}
\textbf{1. General Message Passing (MP):} Let ($K^\ell$, $x^\ell$) denote an attributed simplicial complex at layer $\ell$. TopNets refine the attributed complex using a general TNN layer as 
\begin{align}\label{eq:tnn_topnets}
K^\ell, \tilde{x}^\ell = \mathrm{TNNLayer}_{\ell}(K^{\ell-1},
x^{\ell-1}).
\end{align}

  \textbf{2. PH Vectorization:} Next, we compute a persistence diagram induced by a filtering function $f^\ell$ followed by a vectorization procedure $\psi$. As a result, we obtain a topological vector representation $r_{\sigma}^{\ell}$ for each simplex $\sigma$ in $K^\ell$:
\begin{equation}\label{eq:vectorization_topnets}
    r_{\sigma}^{\ell}  = \psi(\mathrm{PD}(\sigma; f^{\ell}, \tilde{x}^{\ell}, K^{\ell})) \quad \forall \sigma \in K^\ell.
\end{equation}
We note that the map $\mathrm{PD}$ computes persistence diagrams for all dimensions $i=0, 1, \dots, \mathrm{dim}(K^\ell)$.

\vspace{6pt}
\textbf{3. Topological aggregation:}
We combine the PH and MP embeddings of each simplex $\sigma \in K^\ell$ by applying a so-called topological aggregation function $\mathrm{TopAgg}_{\mathrm{dim}(\sigma)}$ --- note that the choice of topological aggregation depends on the dimension of the input simplex. We also group the topological vectors using a dimension-wise $\mathrm{Agg}_{i, \ell}$ operation, i.e.,
\begin{align}
    x^{\ell}_\sigma &= \mathrm{TopAgg}_{\mathrm{dim}(\sigma)}(\tilde{x}^\ell_\sigma, r^\ell_\sigma)  \quad \forall \sigma \in K^\ell \\
    m^{\ell,i} & = \mathrm{Agg}_{i, \ell}(\{r^{\ell}_{\sigma}\}_{\sigma \in K_{[i]}^\ell})
\end{align}


\textbf{4. Readout:} We then merge the features $x^{\ell}_{\sigma}$ and the topological embeddings $m^{\ell, i}$ across layers and, subsequently, across dimensions using interleaved readout functions:
\begin{align}
h^{\ell, i} &= \mathrm{Pool}(\{x_{\sigma}^{\ell}\}_{\sigma \in K_{[i]}^\ell}) \\
h^{i} &= \mathrm{Readout}_{\text{layer}}(\{h^{\ell, i} \}_{\ell}, \{m^{\ell, i}\}_\ell)\\
h &= \mathrm{Readout}_{\text{dim}}(\{h^i\}_i). \label{eq:complex-level-vec}
\end{align}
\end{mybox}

The final representation $h$ in \autoref{eq:complex-level-vec} is typically fed through multi-layer perceptrons (MLP) to obtain a complex-level prediction. Importantly, the formalism of TopNets includes PH-based (graph) neural networks such as TOGL~\citep{horn2021topological}, PersLay~\citep{carrierePersLayNeuralNetwork2020}, and RePHINE~\citep{rephine} as particular cases:

\emph{a) TOGL}: Here, the $\mathrm{TNNLayer}_{\ell}$ functions correspond to GNN layers, while the computation of persistence diagrams ($\mathrm{PD}$) involves vertex-color filtrations, with vectorization achieved via MLPs $\psi$. The topological aggregation $\mathrm{TopAgg}^{\text{TOGL}}$ (defined in Appendix \ref{sec:topnets_deduce}) is specifically applied to persistence tuples of dimension $i=0$, whose vector representations are added to the initial node features. Tuples of dimension $i=1$ are pooled and then concatenated with the final GNN embedding for use in the subsequent readout phase.
\looseness=-1

\emph{b) PersLay}: The $\mathrm{TNNLayer}_{\ell}$ serves as an identity transformation, and the computation of the persistence diagram ($\mathrm{PD}$) involves (0-dim and 1-dim) ordinary and extended persistence pairs. Moreover, $\mathrm{TopAgg}^{\text{PersLay}}$ (defined in Appendix \ref{sec:topnets_deduce}) simply concatenates node features with graph-level topological vectors. 

\emph{c) RePHINE}: Again, GNN is the choice of TNN. However, the computation of persistence diagrams ($\mathrm{PD}$) involves vertex and edge filtrations specific to \emph{RePHINE}. The results are aggregated using a DeepSet function $\mathrm{Agg}_{i,\ell}$ to yield a topological embedding $m^{\ell,i}$ for each layer $\ell$ and dimension $i$. The topological aggregation function  $\mathrm{TopAgg}^{\text{RePHINE}}$ (defined in Appendix \ref{sec:topnets_deduce})outputs the simplex features $\tilde{x}_\sigma^\ell$. Finally, in conjunction with the simplex features from the final layer, the topological embeddings are concatenated and pooled for subsequent use in the downstream readout phase.

More details about deductions can be found in Appendix \ref{sec:topnets_deduce}.

\section{$\mathbf{E(n)}$ Equivariant TopNets}
\label{sec:equiv_ph}
In this section, we extend TopNets to deal with topological objects that are symmetric to rotation, reflections and translations --- i.e., to actions of the Euclidean group $E(n)$. In particular, we consider geometric SCs, and build upon EMPSNs \citep{eijkelboom2023mathrm} and invariant filtering functions to propose Equivariant TopNets (E-TopNets). Compared to regular TopNets, E-TopNets employ modified general message passing and PH vectorization steps (Eqs. \ref{eq:tnn_topnets} and \ref{eq:vectorization_topnets})  --- the other steps remain untouched.  

Starting from an input geometric (attributed) SC  $(K^0, x^0, z^0)$; at each layer $\ell$, E-TopNets recursively obtain a refined SC via an EMPSN layer as
\begin{equation*}
K^{\ell}, \tilde{x}^{\ell}, z^\ell\color{black}= \mathrm{EMPSNLayer}_\ell\color{black}(K^{\ell-1}, x^{\ell-1}, z^{\ell-1}\color{black}).
\end{equation*}
To achieve an equivariant variant of TopNets, one could disregard the vertex coordinates $z^\ell$ when computing persistence diagrams. For instance, this can be obtained from an \emph{$i$-simplex-color filtration} (\autoref{def:i-simplex-color}). This generalizes the notion of vertex-color filtrations to higher dimensions. Thus, $0$-simplex-color filtrations are vertex-color ones, $1$-simplex-color filtrations correspond to edge-color filtrations, and so on.
\looseness=-1

\begin{definition}[$i$-simplex-color filtrations] \label{def:i-simplex-color} Let $(K, x)$ be an attributed simplicial complex and  $f: \mathbb{R}^{d_x} \rightarrow \mathbb{R}^+$ a filtering function. Also, let $\alpha_1 < \dots < \alpha_n$ with $\alpha_j \in \{ f(x_w): w \in K_{[i]}\}$.  An $i$-simplex-color filtration induced by $f$ is a sequence of complexes $K_{\alpha_j} = \{\sigma \in K: o_f(\sigma; x) \leq \alpha_j\}$ for $j=1, \dots, n$, where
$$
o_f(\sigma; x) =\begin{cases} \underset{\tau \subset \sigma: \mathrm{dim}(\tau)=i}{\max}  f(x_\tau)&  \text{, if } \mathrm{dim}(\sigma) \geq i \\
0 & \text{, otherwise}.
\end{cases}
$$
\end{definition}
Obtaining persistence diagrams from $i$-simplex-color filtrations incurs losing (possibly) relevant geometric information. Thus, here, we are interested in filtering functions that leverage both attributes and coordinates, as in geometric color-based filtrations (\autoref{def:geometric-i-simplex-color}).
\begin{definition}[Geometric $i$-simplex-color filtrations] \label{def:geometric-i-simplex-color} Let $(K, x, z)$ be a geometric simplicial complex and $f$ a filtering function. Also, let $\alpha_1 < \dots < \alpha_n$ with $\alpha_j \in \{ f(x_w, \cdot): w \in K_{[i]}\}$.  A geometric $i$-simplex-color filtration induced by $f$ is a sequence $K_{\alpha_j} = \{\sigma \in K: o_f(\sigma; x, z) \leq \alpha_j\}$ for $j=1, \dots, n$, where
$$
o_f(\sigma; x, z) \!=\!\begin{cases} \underset{\substack{\tau \subset \sigma: \\ \mathrm{dim}(\tau)=i}}{\max}  \!\!f(x_\tau, \mathrm{Inv}(\{z_v\}_{v \in \tau}))\!\!\! & \!\!, \mathrm{dim}(\sigma) \geq i \\
0 &\!\! , \text{otherwise}
\end{cases}
$$
and $\mathrm{Inv}(\cdot)$ is any $E(n)$- and $S_n$-invariant function.
\end{definition}


For many tasks, e.g., in graph learning, colors are only given to 0-dim simplices. In such cases, we can obtain colors to higher-order simplices $\sigma$ via a learnable permutation invariant function on the colors of the vertices in $\sigma$. Thus, we can rewrite the filtering functions in \autoref{def:geometric-i-simplex-color} as $f(\phi(\{x_v\}_{v \in \tau}), \mathrm{Inv}(\{z_v\}_{v \in \tau}))$. As usual, we parameterize $f$ using multilayer perceptrons and $\phi$ using DeepSets.  

As a remark, persistence diagrams extracted from geometric $0$-simplex-color filtrations are not more expressive than their non-geometric counterparts --- i.e., vertex-color (VC) filtrations. The reason is that the only $E(n)$-invariant function of a single element is a constant function, i.e., the condition $f(z)=f(\mathfrak{g} \cdot z)$ for all $\mathfrak{g} \in E(n)$ implies that $f$ is a constant function. Thus, we refer to their non-geometric variant whenever we mention VC filtrations.

We also note that, to achieve a geometric extension of RePHINE diagrams, we can simply replace its edge-color filtration with a geometric $1$-simplex-color filtration and then use an independent vertex-color function as in the original formulation. This highlights that the vertex coordinates are only used to define filtrations, and any persistence descriptor and vectorization procedure can be applied --- having no impact on the equivariance of E-TopNets. Our next result (\autoref{prop:invariance_geometric_filtrations}) establishes the invariance of persistence diagrams from geometric $i$-simplex-color filtrations.

\begin{proposition}[Invariant persistence diagrams] \label{prop:invariance_geometric_filtrations}For any $i \geq 0$, persistence diagrams for any dimension obtained from geometric $i$-simplex-color filtrations are $E(n)$-invariant.
\end{proposition}


We can rewrite the PH vectorization step of E-TopNets as
\begin{equation*}
    r_{\sigma}^{\ell}  = \psi(\mathrm{PD}(\sigma; f_{\text{inv}}^{\ell}, \tilde{x}^{\ell}, z^\ell, K^{\ell})) \quad \forall \sigma \in K^\ell
\end{equation*}
where $f_{\text{inv}}^{\ell}$ denotes one or more $E(n)$-invariant filtering functions used to induce a geometric $i$-simplex-color filtration for some $i$ in $\{0, 1, \dots, \mathrm{dim}(K^\ell)\}$.

\section{Continuous (Equivariant) TopNets}
\label{sec:cont_equiv}

In this section, we expand the general framework of (Equivariant) TopNets to encompass continuous systems. Unlike conventional E-TopNets, Continuous E-TopNets use a continuous message-passing scheme based on EMPSNs. For each simplex $\sigma \in K^{t}$ at time-step $t$, we compute the messages $ m^{t, \mathcal{N}}_{\sigma' \rightarrow \sigma} = \mathrm{Msg}_{t, \mathcal{N}}(x^{t}_\sigma, x^{t}_{\sigma'},\mathrm{Inv}(\sigma,\sigma'; z^{t}) )$ from all $\sigma' \in \mathcal{N}(\sigma)$. Then, the messages to simplex $\sigma$ are aggregated using $\mathrm{WithinAgg}_t$ and $\mathrm{BetweenAgg}_t$ the same way as in TNNs to obtain an aggregated message $m^t_\sigma$. Finally, we apply the following functions to obtain the refined feature vectors as 
\begin{align}
    \!\!\dot{\tilde{x}}_\sigma &= \mathrm{Update}(
    m^{t}_\sigma, x^{t}_\sigma)\\
    \!\!\dot{z}_{\sigma} &= C \sum_{\sigma' \in \mathcal{N}_{\uparrow}(\sigma)}(z^{t}_{\sigma} - z^{t}_{\sigma'})\phi_{z}(m^{t, \mathcal{N}_{\uparrow}}_{\sigma' \rightarrow \sigma})~\forall~\sigma~\in~K_{[0]}
\end{align}
where $C$ is a constant and $\phi_{z}$ is an arbitrary non-linear mapping. The forward solution of $x~\text{and}~z$ can be accurately approximated with numerical solvers such as RK4 \citep{runge1895numerische} with low computational cost. The geometrical filtrations and topological embeddings are computed in the same way as described in the previous section.  

Interestingly, one can define a set of associated Neural ODEs for a given PH-based (graph) neural network such as TOGL and RePHINE. We derive the set of neural ODEs and utilize it to derive discretization error bounds between discrete and continuous trajectories.
\begin{table*}[!hbt]
    \caption{\textbf{Predictive performance on graph classification.}}
    \label{tab:graph_clas}
    \centering
    \resizebox{\textwidth}{!}{
    \begin{tabular}{llllcccc cc}
        \toprule
        \textbf{TNN} & \textbf{Topological Agg} & \textbf{Diagram} &\textbf{Method} &\textbf{NCI109} $\uparrow$  &  \textbf{IMDB-B} $\uparrow$   &  \textbf{NCI1} $\uparrow$  &  \textbf{MOLHIV} $\uparrow$   &  \textbf{PROTEINS} $\uparrow$  \\
        \midrule
        \multirow{4}{*}{GCN} &  \multirow{4}{*}{$\mathrm{TopAgg}^{\text{RePHINE}}$}   & \multirow{2}{*}{VC}  & Discrete &  77.92 $\pm$\textcolor{gray}{1.03} &  64.80 $\pm$\textcolor{gray}{1.30} & 79.08 $\pm$\textcolor{gray}{1.06} &  73.64 $\pm$\textcolor{gray}{1.29}& 69.46 $\pm$\textcolor{gray}{1.83} \\
        & &  & Continuous & 80.37 $\pm$\textcolor{gray}{2.21} & 73.40  $\pm$\textcolor{gray}{3.40} & 81.75 $\pm$\textcolor{gray}{2.93} &  72.41 $\pm$\textcolor{gray}{3.29}&  72.89$\pm$\textcolor{gray}{2.10} \\
        & & \multirow{2}{6em}{RePHINE}  & Discrete &  79.18 $\pm$\textcolor{gray}{1.97} &  69.40 $\pm$\textcolor{gray}{3.78} & 80.44 $\pm$\textcolor{gray}{0.94} &  75.98 $\pm$\textcolor{gray}{1.80}& 71.25 $\pm$\textcolor{gray}{1.60} \\
        & &  & Continuous & 80.63 $\pm$\textcolor{gray}{1.56} &  76.00 $\pm$\textcolor{gray}{2.10} &  82.15 $\pm$\textcolor{gray}{1.75} & 74.90 $\pm$\textcolor{gray}{2.78} & 73.79 $\pm$\textcolor{gray}{1.30} \\
        \midrule
        \multirow{4}{*}{GIN} & \multirow{4}{*}{$\mathrm{TopAgg}^{\text{RePHINE}}$} & \multirow{2}{*}{VC}  & Discrete &  78.35 $\pm$\textcolor{gray}{0.68} &  69.80 $\pm$\textcolor{gray}{0.84} & 79.12 $\pm$\textcolor{gray}{1.23} &  73.37 $\pm$\textcolor{gray}{4.36}& 69.46 $\pm$\textcolor{gray}{2.48} \\
        & &  & Continuous & 80.39$\pm$\textcolor{gray}{1.13} &  74.00 $\pm$\textcolor{gray}{3.25} &  82.18 $\pm$\textcolor{gray}{1.56} &  71.90 $\pm$\textcolor{gray}{5.20}& 72.89 $\pm$\textcolor{gray}{2.15} \\
        & & \multirow{2}{*}{RePHINE}  & Discrete &  79.23 $\pm$\textcolor{gray}{1.67} &  72.80 $\pm$\textcolor{gray}{2.95} & 80.92 $\pm$\textcolor{gray}{1.92} &  73.71 $\pm$\textcolor{gray}{0.91}& 72.32 $\pm$\textcolor{gray}{1.89} \\
        & &  & Continuous & 81.60 $\pm$\textcolor{gray}{0.95} &  76.00 $\pm$\textcolor{gray}{1.60}  & 84.16 $\pm$\textcolor{gray}{1.89}& 72.10 $\pm$\textcolor{gray}{4.27}  &  73.79 $\pm$\textcolor{gray}{1.45} \\
        \midrule
        \multirow{4}{*}{MPSN} & \multirow{4}{*}{$\mathrm{TopAgg}^{\text{RePHINE}}$}  & \multirow{2}{*}{VC}  & Discrete &  79.40 $\pm$\textcolor{gray}{2.74} &  66.50 $\pm$\textcolor{gray}{3.65} & 77.10 $\pm$\textcolor{gray}{1.37} &  72.40 $\pm$\textcolor{gray}{3.90}& 70.50 $\pm$\textcolor{gray}{1.75} \\
        & &  & Continuous & 80.10 $\pm$\textcolor{gray}{3.45} &  73.00  $\pm$\textcolor{gray}{1.80} &  81.10 $\pm$\textcolor{gray}{4.64} &  72.70 $\pm$\textcolor{gray}{4.65}&  71.20 $\pm$\textcolor{gray}{3.20} \\
        & & \multirow{2}{*}{RePHINE}  & Discrete &  79.43 $\pm$\textcolor{gray}{1.65} & 67.20   $\pm$\textcolor{gray}{2.85} & 81.22 $\pm$\textcolor{gray}{1.48} &  71.20 $\pm$\textcolor{gray}{4.78}& 71.70 $\pm$\textcolor{gray}{2.56} \\
        & &  & Continuous &  80.40 $\pm$\textcolor{gray}{3.55} &   74.00 $\pm$\textcolor{gray}{2.65}  & 83.20 $\pm$\textcolor{gray}{3.24} & 71.50 $\pm$\textcolor{gray}{4.54} &  72.10 $\pm$\textcolor{gray}{2.35} \\
        \bottomrule
    \end{tabular}}
\end{table*}
\subsection{Discretization Error Bound}
We compute discretization error bounds between the trajectories for discrete and continuous versions of RePHINE and TOGL. All the proofs can be found in Appendix~\ref{sec:bounds}.
\begin{proposition}[Discretization error for TOGL] The discretization error $e_v(\ell)=x_v^{\nicefrac{\ell}{N}} -  x_{v}^\ell$ for node $v$ at layer $\ell$ between the node features of $N$-layer (with time-step size $h$) continuous and discrete TOGL networks is bounded as
\begin{align}
         \|e_v(\ell)\|_1 \leq R_{1}(h)\frac{N(\exp({L_{m} + L_{\beta}}) -1)}{L_{m} + L_{\beta}} 
\end{align}
where $L_m$ and $L_\beta$ are Lipshitz constants, and $R_1$ is a remainder term associated with the Taylor expansion of continuous TOGL.  
\end{proposition}



\begin{proposition}[Discretization error for RePHINE] Let $x_v^{\nicefrac{\ell}{N}}$ and $r^{\nicefrac{\ell}{N}}$ be the node and topological embeddings of an $N$-time-step continuous RePHINE model at time-step $\ell$, respectively. Similarly, let $x_{v}^\ell$ and $r^{\ell}$ be the node and topological embeddings of a discrete $N$-layer RePHINE at layer $\ell$. Then, we can bound the discretization errors $e_{v}(\ell)=x_v^{\nicefrac{\ell}{N}} - x_v^{\ell}$ and $e_{r}(\ell)=r^{\nicefrac{\ell}{N}} - r^{\ell}$ as follows:
\begin{align}
         \|e_{v}(\ell)\|_1 &\leq R_{1}(h)\frac{N(\exp(L_{m}) -1)}{L_{m}}\\
\begin{split}
            \|e_{r}(\ell)\|_1&\leq L^{\ell}_{\beta} \|e_{v}(\ell-1)\|_1+ \frac{L_{\beta}^{\ell}L_{m}}{N} \|e_{v}(\ell-1)\|_1\\ &+ R_{1}(h) - R_{1}(m^{\ell-1})
         \end{split}
\end{align}
where $L_m,L^{\ell}_{\beta}$ are Lipshitz constants,  and $R_1$ are the remainder terms associated with the Taylor expansion of continuous RePHINE.
\end{proposition}

\begin{table}[!htb]
    \caption{\textbf{Comparison with TOGL}. We used $\mathrm{TopAgg}^{\text{TOGL}}$ for aggregating the PH embeddings.}
    \label{tab:togl}
    \centering
    \resizebox{0.48\textwidth}{!}{
    \begin{tabular}{llccc}
        \toprule
        \textbf{Model} &  \textbf{Diagram}  &\textbf{Enzymes} $\uparrow$  &  \textbf{DD} $\uparrow$   &  \textbf{Proteins} $\uparrow$   \\
        \midrule
        GCN & -&   65.8 $\pm$\textcolor{gray}{4.6} &  72.8 $\pm$\textcolor{gray}{4.1} & 76.1 $\pm$\textcolor{gray}{2.4}  \\
        TOGL &  \multirow{2}{*}{VC} &  53.0 $\pm$\textcolor{gray}{9.2} &  73.2 $\pm$\textcolor{gray}{4.7} & 76.0 $\pm$\textcolor{gray}{3.9} \\
        Cont. TopNets & &  69.7 $\pm$\textcolor{gray}{3.2} &  73.1 $\pm$\textcolor{gray}{1.9} &  78.7 $\pm$\textcolor{gray}{2.7}  \\
        \midrule 
        GIN & - &   50.0 $\pm$\textcolor{gray}{12.3} &  70.8 $\pm$\textcolor{gray}{3.8} & 72.3 $\pm$\textcolor{gray}{3.3}  \\
        TOGL & \multirow{2}{*}{VC} &  43.8 $\pm$\textcolor{gray}{7.9} &  75.2 $\pm$\textcolor{gray}{4.2} & 73.6 $\pm$\textcolor{gray}{4.8} \\
        Cont. TopNets & & 58.3 $\pm$\textcolor{gray}{8.2} &  77.3 $\pm$\textcolor{gray}{4.5}  & 79.5 $\pm$\textcolor{gray}{3.9} \\
        \bottomrule
    \end{tabular}}
\end{table}



\textbf{Implication.} The bound indicates that the proximity to the ODE solution cannot be assured since it is uncertain whether $R_{1}(h)N \rightarrow 0$. This suggests the necessity of incorporating additional regulatory assumptions over the network to obtain the Neural ODE in the large depth limit. This observation resonates closely with the analysis conducted by \citet{sander2022residual} in characterizing Neural ODEs with ResNets.

\section{Experiments}

\begin{table*}[!t]
    \caption{\textbf{Test Mean absolute error (MAE) on QM9 dataset}. The $\triangle$ denotes the methods trained with different train-test splits, and $^{**}$ denotes the reproduced results. Benchmarks are from \citet{eijkelboom2023mathrm}. We denote the best-performing methods in \textbf{bold} and the second-best ones in \textcolor{blue}{blue}. We used $\mathrm{TopAgg}^{\text{RePHINE}}$ for aggregating the PH embeddings.}
    \label{tab:qm9_prop}
    \centering
    \resizebox{\textwidth}{!}{
    \begin{tabular}{lcccccc ccccc}
        \toprule
        \textbf{Architecture}  & \textbf{Diagram} &\textbf{Method} &$
        \alpha$  & $\Delta \epsilon$   &  $ \epsilon_{\mathrm{HOMO}}$  & $\epsilon_{\mathrm{LUMO}}$  &  $\mu$ & $C_{v}$ & $R^{2}$ & ZPVE  \\
        &  & & $\mathrm{bohr}^{3}$ & meV & meV & meV & D & cal/mol K & $\mathrm{bohr}^{3}$ & meV\\
        \midrule
        DimeNet$++^{\triangle}$& - & - &0.044 & 33   & 25  &20  &  0.030 & 0.023 & 0.331 & 1.21\\
        SphereNet$^{\triangle}$&  - &- &0.046 & 32   & 23  &18  &  0.026 & 0.021 & 0.292 & 1.21\\
        \midrule
        NMP &  - &- &0.092 & 69   & 43  &38  &  0.030 & 0.040 & 0.180 & 1.50  \\
       $\mathrm{SE}(3)$-Tr &  - &- &0.142 & 53   & 35  &33  &  0.051 & 0.054 & - & -  \\
        TFN & - & -&0.223 & 58   & 40  &38  &  0.064 & 0.101 & - & -  \\
        MPSN & - & - &0.266 & 153   & 89  &77  &  0.101 & 0.122 & 0.887 & 3.02  \\
        EGNN & - &- &0.071 & \textcolor{blue}{48}   & \textbf{29 } &25  &  \textbf{0.028} & 0.031 & \textbf{0.106} & 1.55  \\
        IMPSN$^{**}$ &  - &- & \textbf{0.066} &  51 &  \textcolor{blue}{32} & 25 &  0.031 & \textbf{0.027} &  \textcolor{blue}{0.114} & 1.44  \\
        \midrule
        \multirow{4}{*}{IMPSN} & \multirow{2}{*}{VC}  & Disc. E-TopNets & 0.083 & \textbf{47}   & 37  & \textbf{24} & 0.035  & 0.032  & 0.125  & 1.45  \\
        &   & Cont. E-TopNets & 0.075 & 49  &  36 & 27 &  0.030 & 0.035  & 0.129 & 1.43   \\
        &  \multirow{2}{*}{RePHINE}  & Disc. E-TopNets & 0.072 &  57  &  33 & 28 &  \textcolor{blue}{0.029} &  \textcolor{blue}{0.028} & 0.132 & \textcolor{blue}{1.39}  \\
        &   & Cont. E-TopNets &\textcolor{blue}{0.070}  & 50 & 35  & 25 & 0.032 & 0.030 & 0.118 &  \textbf{1.37} \\
        \bottomrule
    \end{tabular}}
    \vspace{-8pt}
\end{table*}

\textbf{Tasks.}
We assess the performance of TopNets on diverse tasks: (i) we evaluate our method performance on real-world graph classification data while considering discrete and continuous versions of various GNNs and TNNs in Section~\ref{subsec:graph_classif}, (ii) we benchmark TopNets efficacy in property prediction using QM9 molecular data, highlighting the effectiveness of its equivariant variant in Section~\ref{subsec:equiv_pred}, (iii) we demonstrate TopNets utility in co-designing antibody sequence and structure using the SAbDab database in Section~\ref{subsec:cdrh3}, and (iv) we evaluate our method on 3BPA MD17 trajectories \citep{kovacs2021linear} in Section \ref{subsec:md17}. 

\textbf{Baselines.} On graph classification tasks, we use standard vertex-color (VC) and RePHINE \citep{rephine} to compute persistence diagrams. We adopt different GNN/TNN architectures like GCN \citep{kipf2016semi}, GIN \citep{gin}, TOGL \citep{horn2021topological}, and MPSN \citep{Bodnar2021}. We also compare the performance between each method's continuous and discrete counterparts. On QM9 property prediction tasks, we compare TopNets to several equivariant methods like NMP \citep{gilmer2017neural}, TFN \citep{thomas2018tensor}, $\mathrm{SE}(3)$-Tr \citep{fuchs2020se}, DimeNet++ \citep{gasteiger2020fast}, SphereNet \citep{liu2021spherical}, MPSN\citep{Bodnar2021}, EGNN \citep{satorras2021n} and IMPSN \citep{eijkelboom2023mathrm}. On CDR-H3 Antibody design, we compare to recent SOTA like RefineGNN~\citep{refinegnn}, MEAN~\citep{mean} and AbODE~\citep{pmlr-v202-verma23a}. Lastly, for 3BPA MD17 trajectories, we compare our method to SOTA like NequIP \cite{Batzner_2022} and MACE \citep{batatia2022mace} which use higher-order message passing mechanisms. 
\looseness=-1

Implementation details are given in Appendix~\ref{sec:implement}.


\subsection{Graph Classification}
\label{subsec:graph_classif}

The results presented in \autoref{tab:graph_clas} and \autoref{tab:togl} demonstrate the performance of TopNets on graph classification. These results offer a detailed assessment of different GNN/TNN architectures, PH vectorization methods, and their continuous counterparts. The reported results include the mean and standard deviation of predictive metrics --- AUROC for MOLHIV and accuracy for the remaining datasets. This comprehensive analysis provides valuable insights into TopNets performance. Notably, incorporating the continuous component consistently improves downstream performance across all datasets, TNNs, and $\mathrm{TopAgg}$ schemes.


\subsection{Molecular data - QM9}
\label{subsec:equiv_pred}
The QM9 dataset, introduced by \citet{ramakrishnan2014quantum}, comprises small molecules with a maximum of 29 atoms in 3D space. Each atom is characterized by a 3D position and a five-dimensional one-hot node embedding representing the atom type, denoted as $(\mathrm{H}, \mathrm{C}, \mathrm{N}, \mathrm{O}, \mathrm{F})$. The dataset's primary objective is to predict various chemical properties of the molecules, which remain invariant to translations, rotations, and reflections on the atom positions. Following the data preparation strategy of \citet{eijkelboom2023mathrm, satorras2021n}, we partition the dataset into training, validation, and test sets. The mean absolute error between predictions and ground truth for test set is reported in \autoref{tab:qm9_prop}, revealing the competitive performance of TopNets compared to baselines. Notably, on many targets, TopNets achieve results nearly on par with SOTA approaches, surpassing in predicting ZPVE, $\Delta \epsilon$ and $\epsilon_{LUMO}$. This achievement is intriguing as our architecture, not specifically tailored for molecular tasks, lacks many molecule-specific intricacies, like Bessel function embeddings \citep{gasteiger2020directional}.

\subsection{CDR-H3 Antibody Design}
\label{subsec:cdrh3}
We took the antigen-antibody complexes dataset from Structural Antibody Database~\cite{dunbar2014sabdab} and removed invalid data points. We followed a strategy similar to \citet{pmlr-v202-verma23a} for data preparation and splitting and employ Amino Acid Recovery (AAR) and RMSD for quantitative evaluation. AAR is defined as the overlapping rate between the predicted 1D sequences and the ground truth. RMSD is calculated via the Kabsch algorithm~\cite{kabsch1976solution} based on $C_{\alpha}$ spatial features of the CDR residues.
\begin{table}[!hbt]
\centering
\caption{ Results on CDR-H3 design benchmark. We report AAR and RMSD metrics. TopNets significantly outperform baselines on AAR while being competitive on RMSD. We used $\mathrm{TopAgg}^{\text{RePHINE}}$ for aggregating the PH embeddings.}
\label{tab:result_cdr3}
\centering
\resizebox{0.46\textwidth}{!}{
\begin{tabular}{lccc}
\toprule
Method & Diagram &AAR $\%$ ($\uparrow$) & RMSD ($\downarrow$)   \\
\midrule
LSTM & -&15.69 $\pm$ \textcolor{gray}{0.91} & \textcolor{gray}{(N/A)} \\
C-LSTM & -&15.48 $\pm$ \textcolor{gray}{1.17} & \textcolor{gray}{(N/A)}  \\
RefineGNN & -&21.13 $\pm$ \textcolor{gray}{1.59}& 6.00 $\pm$ \textcolor{gray}{0.55}   \\
C-RefineGNN & -&18.88 $\pm$ \textcolor{gray}{1.37} & 6.22 $\pm$ \textcolor{gray}{0.59}   \\
MEAN & -&36.38 $\pm$ \textcolor{gray}{3.08} & 2.21 $\pm$ \textcolor{gray}{0.16}  \\
AbODE &  -&39.8 $\pm$ \textcolor{gray}{1.17} & 1.73 $\pm$ \textcolor{gray}{0.11}    \\
\midrule
\multirow{2}{4em}{TopNets} &  VC& 43.00 $\pm$ \textcolor{gray}{1.34} & 1.73 $\pm$ \textcolor{gray}{0.21}    \\
&  RePHINE&\textbf{44.80} $\pm$ \textcolor{gray}{1.57} & 1.75 $\pm$ \textcolor{gray}{0.17}    \\
\hline
\end{tabular}}
\end{table}

\autoref{tab:result_cdr3} showcases the performance of TopNets compared to the baseline methods over CDR-H3 design. TopNets outperform other methods in terms of sequence prediction, thus improving over the SOTA and demonstrating the benefit of persistent homology in generative design.


\subsection{Molecular Dynamics}
\label{subsec:md17}

We use the standard 3BPA MD17 \citep{kovacs2021linear} dataset  to evaluate the extrapolation capabilities. The training set consists of $500$ geometries sampled from $300$ K molecular dynamics simulation of the large and flexible drug-like molecule 3-(benzyloxy)pyridin-2-amine. The three test sets contain geometries sampled at 300 K, 600 K, and 1200 K to assess in- and out-of-domain accuracy. The task is to predict the (E, meV) and force (F, meV/$\mathring{A}$) of the given conformations. In order to have a fair comparison, we followed the same data-preparation strategy and training setup as described in \citet{batatia2022mace}. 
\begin{table}[!hbt]
\centering
    \caption{RMSE on 3BPA MD17 dataset.}
    \label{tab:md17}
    \resizebox{0.47\textwidth}{!}{
    \begin{tabular}{lccccc}
        \toprule
        Task & Variable &NequIP & MACE &TopNets \\
        \midrule
        \multirow{2}{*}{300K} &  E  & 3.3$\pm$\textcolor{gray}{0.1} & 3.0$\pm$\textcolor{gray}{0.2} & \textbf{2.5}$\pm$\textcolor{gray}{0.2}\\
        & F & 10.8$\pm$\textcolor{gray}{0.2} & \textbf{8.8}$\pm$\textcolor{gray}{0.3} &8.9 $\pm$\textcolor{gray}{0.2} \\
        \midrule
        \multirow{2}{*}{600K} &  E   & 11.2$\pm$\textcolor{gray}{0.1}  & 9.7$\pm$\textcolor{gray}{0.5} & \textbf{9.5}$\pm$\textcolor{gray}{0.7} \\
        & F & 26.4$\pm$\textcolor{gray}{0.1}  & 21.8$\pm$\textcolor{gray}{0.6} &21.8 $\pm$\textcolor{gray}{0.7} \\
        \midrule
        \multirow{2}{*}{1200K} & E   &  38.5$\pm$\textcolor{gray}{1.6} & 29.8$\pm$\textcolor{gray}{1.0} &\textbf{29.2}$\pm$\textcolor{gray}{1.2} \\
        & F & 76.2$\pm$\textcolor{gray}{1.1} & \textbf{62.0} $\pm$\textcolor{gray}{1.8} &62.3 $\pm$\textcolor{gray}{0.5} \\
        \bottomrule
    \end{tabular}}
    \vspace{-4pt}
\end{table}

\autoref{tab:md17} showcases the performance of TopNets compared to the baseline methods over energy (E, meV) and force (F, meV/$\mathring{A}$) on different sets of geometries sampled at 300 K, 600 K, and 1200 K. We used $\mathrm{TopAgg}^{\mathrm{RePHINE}}$ to aggregate and compute the PH embeddings. Notably, our results indicate that TopNets and MACE generally outperform NequIP in predicting energy as well as force of the molecular configurations. 
\looseness=-1


\section{Ablations}

\textbf{Runtime comparison.} We conducted an ablation study to characterize the runtime complexity of our method, assessing the time taken per epoch to train different models on a single V100 GPU. The results are shown in \autoref{tab:runtime}, and as expected, continuous methods require additional time due to solving the ODE forward compared to their discrete counterparts. However, the ODE-based methods can be trained without storing intermediate quantities, leading to a  $\mathcal{O}(1)$ constant memory requirement \citep{ODE} as compared to the memory cost of training other methods which increases with the depth of the network. 

\textbf{Higher-order PH.}
We conducted an ablation study to evaluate the influence of higher-order persistent homology (PH) features on TNNs. We employed fixed filtering functions based on curvature filtrations \citep{southern2024curvature} to extract $\mathcal{D}_0,\mathcal{D}_1,\mathcal{D}_2$ level diagrams corresponding to $0,1,2$-dim topological features. These features are then aggregated using deepsets \citep{deepsets} to combine with pooled node features extracted by the TNN for downstream tasks. We evaluate the method on two graph classification datasets: IMDB-Binary and Proteins --- we apply uplifting to obtain clique complexes from which we extract persistence diagrams \citep{southern2024curvature}. The results shown in \autoref{tab:high_ph} demonstrate that incorporating higher-order PH features leads to improved performance across various TNNs on both datasets.
\begin{table}[!hbt]
    \centering
    \caption{Runtime comparison. We use $\mathrm{TopAgg}^{\text{RePHINE}}$ as topological aggregator.}
    \label{tab:runtime}
    \resizebox{0.45\textwidth}{!}{
    \begin{tabular}{llllc}
        \toprule
        TNN &  Diagram& Method & Time (GPU time)\\
        \midrule
        \multirow{4}{*}{GCN}  & \multirow{2}{*}{VC}  & Discrete &  62.9 $\pm$\textcolor{gray}{11.5} \\
        &   & Continuous & 106.8 $\pm$\textcolor{gray}{12.8} \\
        &  \multirow{2}{6em}{RePHINE}  & Discrete &  62.7 $\pm$\textcolor{gray}{13.4}  \\
        &   & Continuous &  112.7 $\pm$\textcolor{gray}{16.7} \\
        \midrule
        \multirow{4}{*}{GIN} & \multirow{2}{*}{VC}  & Discrete &  60.7 $\pm$\textcolor{gray}{8.7}  \\
        &   & Continuous & 105.7 $\pm$\textcolor{gray}{10.8} \\
         & \multirow{2}{*}{RePHINE}  & Discrete &  65.9 $\pm$\textcolor{gray}{10.5} \\
         &  & Continuous &  120.7 $\pm$\textcolor{gray}{15.7} \\
        \bottomrule
    \end{tabular}}
\end{table}

\begin{table}[!hbt]
    \caption{Predictive graph classification when including higher-order PH features.}
    \label{tab:high_ph}
    \centering
    \resizebox{0.3\textwidth}{!}{
    \begin{tabular}{llccc}
        \toprule
        TNN & IMDB-B $\uparrow$ & Proteins $\uparrow$ \\
        \midrule
        GCN &  73.0 $\pm$\textcolor{gray}{1.30}&  71.4 $\pm$\textcolor{gray}{1.10}\\
        GIN &   76.0 $\pm$\textcolor{gray}{1.70} &  74.2 $\pm$\textcolor{gray}{2.10} \\
        MPSN &  76.0 $\pm$\textcolor{gray}{1.50} &  74.5 $\pm$\textcolor{gray}{2.75} \\
        \bottomrule
    \end{tabular}}
\end{table}

\section{Conclusion and Limitations}
We introduce TopNets to illustrate the theoretical and practical benefits of including persistent features in topological networks, and their geometric and continuous-time extensions. 
TopNets incur considerable computational expense due to costs involved in computing PH embeddings as well as higher-order message-passing. Additionally, our research is confined to simplicial complexes, and exploring combinatorial complexes is an interesting avenue for future work.

\section*{Acknowledgements}
This work has been supported by the Research Council of Finland under the {\em HEALED} project (grant 13342077), Jane and Aatos Erkko Foundation  project (grant 7001703) on ``Biodesign: Use of artificial intelligence in enzyme design for synthetic biology'', and  Finnish Center for Artificial Intelligence FCAI (Flagship programme). We acknowledge CSC – IT Center for Science, Finland, for providing generous computational resources. 

\section*{Impact Statement}
We proposed a framework for topological representation learning, leveraging persistent homology and topological neural networks. Our contributions advance current art by offering a unified framework for topological deep learning, and extending it to accommodate geometric features naturally appearing in many downstream real-world applications. Potential applications in drug property prediction, molecular simulation, and protein generative design have significant implications for drug development and precision medicine.

\bibliography{sample}

\begin{thebibliography}{72}
\providecommand{\natexlab}[1]{#1}
\providecommand{\url}[1]{\texttt{#1}}
\expandafter\ifx\csname urlstyle\endcsname\relax
  \providecommand{\doi}[1]{doi: #1}\else
  \providecommand{\doi}{doi: \begingroup \urlstyle{rm}\Url}\fi

\bibitem[Adams et~al.(2016)Adams, Emerson, Kirby, Neville, Peterson, Shipman, Chepushtanova, Hanson, Motta, and Ziegelmeier]{adamsPersistenceImagesStable2016}
Henry Adams, Tegan Emerson, Michael Kirby, Rachel Neville, Chris Peterson, Patrick Shipman, Sofya Chepushtanova, Eric Hanson, Francis Motta, and Lori Ziegelmeier.
\newblock Persistence {{Images}}: {{A Stable Vector Representation}} of {{Persistent Homology}}.
\newblock \emph{Journal of Machine Learning Research}, 18\penalty0 (8):\penalty0 1--35, 2016.

\bibitem[Barbarossa and Sardellitti(2020)]{barbarossa2020topological}
Sergio Barbarossa and Stefania Sardellitti.
\newblock Topological signal processing over simplicial complexes.
\newblock \emph{IEEE Transactions on Signal Processing}, 68:\penalty0 2992--3007, 2020.

\bibitem[Batatia et~al.(2022)Batatia, Kovacs, Simm, Ortner, and Cs{\'a}nyi]{batatia2022mace}
Ilyes Batatia, David~P Kovacs, Gregor Simm, Christoph Ortner, and G{\'a}bor Cs{\'a}nyi.
\newblock Mace: Higher order equivariant message passing neural networks for fast and accurate force fields.
\newblock \emph{Advances in Neural Information Processing Systems}, 35:\penalty0 11423--11436, 2022.

\bibitem[Batzner et~al.(2022)Batzner, Musaelian, Sun, Geiger, Mailoa, Kornbluth, Molinari, Smidt, and Kozinsky]{Batzner_2022}
Simon Batzner, Albert Musaelian, Lixin Sun, Mario Geiger, Jonathan~P. Mailoa, Mordechai Kornbluth, Nicola Molinari, Tess~E. Smidt, and Boris Kozinsky.
\newblock E(3)-equivariant graph neural networks for data-efficient and accurate interatomic potentials.
\newblock \emph{Nature Communications}, 13\penalty0 (1), May 2022.
\newblock ISSN 2041-1723.
\newblock \doi{10.1038/s41467-022-29939-5}.

\bibitem[Bodnar et~al.(2021{\natexlab{a}})Bodnar, Frasca, Otter, Wang, Li\`{o}, Montufar, and Bronstein]{Bodnar2021}
Cristian Bodnar, Fabrizio Frasca, Nina Otter, Yuguang Wang, Pietro Li\`{o}, Guido~F Montufar, and Michael Bronstein.
\newblock Weisfeiler and lehman go cellular: Cw networks.
\newblock In \emph{Advances in Neural Information Processing Systems (NeurIPS)}, 2021{\natexlab{a}}.

\bibitem[Bodnar et~al.(2021{\natexlab{b}})Bodnar, Frasca, Wang, Otter, Montufar, Lio, and Bronstein]{bodnar2021weisfeiler}
Cristian Bodnar, Fabrizio Frasca, Yuguang Wang, Nina Otter, Guido~F Montufar, Pietro Lio, and Michael Bronstein.
\newblock Weisfeiler and lehman go topological: Message passing simplicial networks.
\newblock In \emph{International Conference on Machine Learning (ICML)}, 2021{\natexlab{b}}.

\bibitem[Brandstetter et~al.(2023)Brandstetter, Berg, Welling, and Gupta]{brandstetter2022clifford}
Johannes Brandstetter, Rianne van~den Berg, Max Welling, and Jayesh Gupta.
\newblock Clifford neural layers for {PDE} modeling.
\newblock In \emph{ICLR}, 2023.

\bibitem[Brehmer et~al.(2023)Brehmer, De~Haan, Behrends, and Cohen]{brehmer2023geometric}
Johann Brehmer, Pim De~Haan, S{\"o}nke Behrends, and Taco Cohen.
\newblock Geometric algebra transformers.
\newblock \emph{arXiv preprint arXiv:2305.18415}, 2023.

\bibitem[Bronstein et~al.(2021)Bronstein, Bruna, Cohen, and Velicković]{bronstein2021geometric}
Michael~M. Bronstein, Joan Bruna, Taco Cohen, and Petar Velicković.
\newblock Geometric deep learning: Grids, groups, graphs, geodesics, and gauges, 2021.

\bibitem[Bubenik(2015)]{bubenikStatisticalTopologicalData2015}
P.~Bubenik.
\newblock Statistical topological data analysis using persistence landscapes.
\newblock \emph{Journal of Machine Learning Research}, 16:\penalty0 77--102, 2015.

\bibitem[Carri{\`e}re et~al.(2020)Carri{\`e}re, Chazal, Ike, Lacombe, Royer, and Umeda]{carrierePersLayNeuralNetwork2020}
Mathieu Carri{\`e}re, Fr{\'e}d{\'e}ric Chazal, Yuichi Ike, Th{\'e}o Lacombe, Martin Royer, and Yuhei Umeda.
\newblock {{PersLay}}: {{A Neural Network Layer}} for {{Persistence Diagrams}} and {{New Graph Topological Signatures}}.
\newblock In \emph{Artificial Intelligence and Statistics (AISTATS)}, 2020.

\bibitem[Chamberlain et~al.(2021)Chamberlain, Rowbottom, Gorinova, Bronstein, Webb, and Rossi]{chamberlain2021grand}
Ben Chamberlain, James Rowbottom, Maria~I Gorinova, Michael Bronstein, Stefan Webb, and Emanuele Rossi.
\newblock Grand: Graph neural diffusion.
\newblock In \emph{International Conference on Machine Learning}, pages 1407--1418. PMLR, 2021.

\bibitem[Chen et~al.(2019)Chen, Deng, and Hu]{chen2019mixed}
Binghui Chen, Weihong Deng, and Jiani Hu.
\newblock Mixed high-order attention network for person re-identification.
\newblock In \emph{Proceedings of the IEEE/CVF international conference on computer vision}, pages 371--381, 2019.

\bibitem[Chen et~al.(2018)Chen, Rubanova, Bettencourt, and Duvenaud]{ODE}
Ricky T.~Q. Chen, Yulia Rubanova, Jesse Bettencourt, and David~K Duvenaud.
\newblock Neural ordinary differential equations.
\newblock In S.~Bengio, H.~Wallach, H.~Larochelle, K.~Grauman, N.~Cesa-Bianchi, and R.~Garnett, editors, \emph{Advances in Neural Information Processing Systems}, volume~31. Curran Associates, Inc., 2018.

\bibitem[Choi et~al.(2022)Choi, Hong, Park, and Cho]{choi2022gread}
Jeongwhan Choi, Seoyoung Hong, Noseong Park, and Sung-Bae Cho.
\newblock Gread: Graph neural reaction-diffusion equations.
\newblock \emph{arXiv preprint arXiv:2211.14208}, 2022.

\bibitem[Demailly(2006)]{demailly2006analyse}
Jean-Pierre Demailly.
\newblock \emph{Analyse num{\'e}rique et {\'e}quations diff{\'e}rentielles}.
\newblock EDP sciences Les Ulis, 2006.

\bibitem[Dong et~al.(2020)Dong, Sawin, and Bengio]{dong2020hnhn}
Yihe Dong, Will Sawin, and Yoshua Bengio.
\newblock Hnhn: Hypergraph networks with hyperedge neurons.
\newblock \emph{arXiv preprint arXiv:2006.12278}, 2020.

\bibitem[Dunbar et~al.(2014)Dunbar, Krawczyk, Leem, Baker, Fuchs, Georges, Shi, and Deane]{dunbar2014sabdab}
James Dunbar, Konrad Krawczyk, Jinwoo Leem, Terry Baker, Angelika Fuchs, Guy Georges, Jiye Shi, and Charlotte~M Deane.
\newblock Sabdab: the structural antibody database.
\newblock \emph{Nucleic acids research}, 42\penalty0 (D1):\penalty0 D1140--D1146, 2014.

\bibitem[Dupont et~al.(2019)Dupont, Doucet, and Teh]{ode2}
Emilien Dupont, Arnaud Doucet, and Yee~Whye Teh.
\newblock Augmented neural odes.
\newblock \emph{Advances in neural information processing systems}, 32, 2019.

\bibitem[Duval et~al.(2023)Duval, Mathis, Joshi, Schmidt, Miret, Malliaros, Cohen, Lio, Bengio, and Bronstein]{duval2023hitchhikers}
Alexandre Duval, Simon~V. Mathis, Chaitanya~K. Joshi, Victor Schmidt, Santiago Miret, Fragkiskos~D. Malliaros, Taco Cohen, Pietro Lio, Yoshua Bengio, and Michael Bronstein.
\newblock A hitchhiker's guide to geometric gnns for 3d atomic systems, 2023.

\bibitem[Edelsbrunner and Harer(2010)]{ComputationalTopoBook}
H.~Edelsbrunner and J.~Harer.
\newblock \emph{Computational Topology - an Introduction.}
\newblock American Mathematical Society, 2010.

\bibitem[Eijkelboom et~al.(2023)Eijkelboom, Hesselink, and Bekkers]{eijkelboom2023mathrm}
Floor Eijkelboom, Rob Hesselink, and Erik Bekkers.
\newblock E(n) equivariant message passing simplicial networks.
\newblock \emph{arXiv preprint arXiv:2305.07100}, 2023.

\bibitem[Freeman(2004)]{freeman2004development}
Linton Freeman.
\newblock The development of social network analysis.
\newblock \emph{A Study in the Sociology of Science}, 1\penalty0 (687):\penalty0 159--167, 2004.

\bibitem[Fuchs et~al.(2020)Fuchs, Worrall, Fischer, and Welling]{fuchs2020se}
Fabian Fuchs, Daniel Worrall, Volker Fischer, and Max Welling.
\newblock Se (3)-transformers: 3d roto-translation equivariant attention networks.
\newblock \emph{Advances in neural information processing systems}, 33:\penalty0 1970--1981, 2020.

\bibitem[Garg et~al.(2020)Garg, Jegelka, and Jaakkola]{gnns_repr}
Vikas~K. Garg, Stefanie Jegelka, and Tommi Jaakkola.
\newblock Generalization and representational limits of graph neural networks.
\newblock In \emph{International Conference on Machine Learning}, 2020.

\bibitem[Gasteiger et~al.(2020{\natexlab{a}})Gasteiger, Giri, Margraf, and G{\"u}nnemann]{gasteiger2020fast}
Johannes Gasteiger, Shankari Giri, Johannes~T Margraf, and Stephan G{\"u}nnemann.
\newblock Fast and uncertainty-aware directional message passing for non-equilibrium molecules.
\newblock \emph{arXiv preprint arXiv:2011.14115}, 2020{\natexlab{a}}.

\bibitem[Gasteiger et~al.(2020{\natexlab{b}})Gasteiger, Gro{\ss}, and G{\"u}nnemann]{gasteiger2020directional}
Johannes Gasteiger, Janek Gro{\ss}, and Stephan G{\"u}nnemann.
\newblock Directional message passing for molecular graphs.
\newblock \emph{arXiv preprint arXiv:2003.03123}, 2020{\natexlab{b}}.

\bibitem[Geiger and Smidt(2022)]{geiger2022e3nn}
Mario Geiger and Tess Smidt.
\newblock e3nn: Euclidean neural networks.
\newblock \emph{arXiv preprint arXiv:2207.09453}, 2022.

\bibitem[Gilmer et~al.(2017)Gilmer, Schoenholz, Riley, Vinyals, and Dahl]{gilmer2017neural}
Justin Gilmer, Samuel~S Schoenholz, Patrick~F Riley, Oriol Vinyals, and George~E Dahl.
\newblock Neural message passing for quantum chemistry.
\newblock In \emph{International conference on machine learning}, pages 1263--1272. PMLR, 2017.

\bibitem[Giusti et~al.(2023)Giusti, Battiloro, Testa, Di~Lorenzo, Sardellitti, and Barbarossa]{giusti2023cell}
Lorenzo Giusti, Claudio Battiloro, Lucia Testa, Paolo Di~Lorenzo, Stefania Sardellitti, and Sergio Barbarossa.
\newblock Cell attention networks.
\newblock In \emph{2023 International Joint Conference on Neural Networks (IJCNN)}, pages 1--8. IEEE, 2023.

\bibitem[Grathwohl et~al.(2018)Grathwohl, Chen, Bettencourt, Sutskever, and Duvenaud]{grathwohl2018ffjord}
Will Grathwohl, Ricky~TQ Chen, Jesse Bettencourt, Ilya Sutskever, and David Duvenaud.
\newblock Ffjord: Free-form continuous dynamics for scalable reversible generative models.
\newblock \emph{arXiv preprint arXiv:1810.01367}, 2018.

\bibitem[Hensel et~al.(2021)Hensel, Moor, and Rieck]{Hensel21}
Felix Hensel, Michael Moor, and Bastian Rieck.
\newblock A survey of topological machine learning methods.
\newblock \emph{Frontiers in Artificial Intelligence}, 4, 2021.

\bibitem[Hofer et~al.(2017)Hofer, Kwitt, Niethammer, and Uhl]{Hofer2017}
C.~Hofer, R.~Kwitt, M.~Niethammer, and A.~Uhl.
\newblock Deep learning with topological signatures.
\newblock In \emph{Advances in Neural Information Processing Systems (NeurIPS)}, 2017.

\bibitem[Horn et~al.(2021)Horn, De~Brouwer, Moor, Moreau, Rieck, and Borgwardt]{horn2021topological}
Max Horn, Edward De~Brouwer, Michael Moor, Yves Moreau, Bastian Rieck, and Karsten Borgwardt.
\newblock Topological graph neural networks.
\newblock \emph{arXiv preprint arXiv:2102.07835}, 2021.

\bibitem[Iakovlev et~al.(2020)Iakovlev, Heinonen, and L{\"a}hdesm{\"a}ki]{iakovlev2020learning}
Valerii Iakovlev, Markus Heinonen, and Harri L{\"a}hdesm{\"a}ki.
\newblock Learning continuous-time pdes from sparse data with graph neural networks.
\newblock \emph{arXiv preprint arXiv:2006.08956}, 2020.

\bibitem[Immonen et~al.(2023)Immonen, Souza, and Garg]{rephine}
Johanna Immonen, Amauri~H. Souza, and Vikas Garg.
\newblock Going beyond persistent homology using persistent homology.
\newblock In \emph{Advances in Neural Information Processing Systems (NeurIPS)}, 2023.

\bibitem[Jha et~al.(2022)Jha, Saha, and Singh]{jha2022prediction}
Kanchan Jha, Sriparna Saha, and Hiteshi Singh.
\newblock Prediction of protein--protein interaction using graph neural networks.
\newblock \emph{Scientific Reports}, 12\penalty0 (1):\penalty0 8360, 2022.

\bibitem[Jin et~al.(2022)Jin, Wohlwend, Barzilay, and Jaakkola]{refinegnn}
Wengong Jin, Jeremy Wohlwend, Regina Barzilay, and Tommi Jaakkola.
\newblock Iterative refinement graph neural network for antibody sequence-structure co-design, 2022.

\bibitem[Joshi et~al.(2023)Joshi, Bodnar, Mathis, Cohen, and Liò]{joshi2023expressive}
Chaitanya~K. Joshi, Cristian Bodnar, Simon~V. Mathis, Taco Cohen, and Pietro Liò.
\newblock On the expressive power of geometric graph neural networks, 2023.

\bibitem[Kabsch(1976)]{kabsch1976solution}
Wolfgang Kabsch.
\newblock A solution for the best rotation to relate two sets of vectors.
\newblock \emph{Acta Crystallographica Section A: Crystal Physics, Diffraction, Theoretical and General Crystallography}, 32\penalty0 (5):\penalty0 922--923, 1976.

\bibitem[Kim et~al.(2023)Kim, Can, and Krishnamurthy]{kim2023trainability}
Timothy~Doyeon Kim, Tankut Can, and Kamesh Krishnamurthy.
\newblock Trainability, expressivity and interpretability in gated neural odes.
\newblock \emph{arXiv preprint arXiv:2307.06398}, 2023.

\bibitem[Kipf and Welling(2016)]{kipf2016semi}
Thomas~N Kipf and Max Welling.
\newblock Semi-supervised classification with graph convolutional networks.
\newblock \emph{arXiv preprint arXiv:1609.02907}, 2016.

\bibitem[Kochkov et~al.(2021)Kochkov, Smith, Alieva, Wang, Brenner, and Hoyer]{kochkov2021machine}
Dmitrii Kochkov, Jamie Smith, Ayya Alieva, Qing Wang, Michael Brenner, and Stephan Hoyer.
\newblock Machine learning--accelerated computational fluid dynamics.
\newblock \emph{Proceedings of the National Academy of Sciences}, 118\penalty0 (21), 2021.

\bibitem[Kong et~al.(2023)Kong, Huang, and Liu]{mean}
Xiangzhe Kong, Wenbing Huang, and Yang Liu.
\newblock Conditional antibody design as 3d equivariant graph translation, 2023.

\bibitem[Kov{\'a}cs et~al.(2021)Kov{\'a}cs, Oord, Kucera, Allen, Cole, Ortner, and Cs{\'a}nyi]{kovacs2021linear}
D{\'a}vid~P{\'e}ter Kov{\'a}cs, Cas van~der Oord, Jiri Kucera, Alice~EA Allen, Daniel~J Cole, Christoph Ortner, and G{\'a}bor Cs{\'a}nyi.
\newblock Linear atomic cluster expansion force fields for organic molecules: beyond rmse.
\newblock \emph{Journal of chemical theory and computation}, 17\penalty0 (12):\penalty0 7696--7711, 2021.

\bibitem[Kusano et~al.(2016)Kusano, Hiraoka, and Fukumizu]{kusanoPersistenceWeightedGaussian2016}
Genki Kusano, Yasuaki Hiraoka, and Kenji Fukumizu.
\newblock Persistence weighted {{Gaussian}} kernel for topological data analysis.
\newblock In \emph{{{International Conference}} on {{Machine Learning}} (ICML)}, pages 2004--2013, 2016.

\bibitem[Li et~al.(2021)Li, Kovachki, Azizzadenesheli, Liu, Bhattacharya, Stuart, and Anandkumar]{li2020fourier}
Zongyi Li, Nikola Kovachki, Kamyar Azizzadenesheli, Burigede Liu, Kaushik Bhattacharya, Andrew Stuart, and Anima Anandkumar.
\newblock Fourier neural operator for parametric partial differential equations.
\newblock In \emph{ICLR}, 2021.

\bibitem[Lipman et~al.(2023)Lipman, Chen, Ben-Hamu, Nickel, and Le]{lipman2023flow}
Yaron Lipman, Ricky T.~Q. Chen, Heli Ben-Hamu, Maximilian Nickel, and Matt Le.
\newblock Flow matching for generative modeling, 2023.

\bibitem[Liu et~al.(2021)Liu, Wang, Liu, Zhang, Oztekin, and Ji]{liu2021spherical}
Yi~Liu, Limei Wang, Meng Liu, Xuan Zhang, Bora Oztekin, and Shuiwang Ji.
\newblock Spherical message passing for 3d graph networks.
\newblock \emph{arXiv preprint arXiv:2102.05013}, 2021.

\bibitem[Lu et~al.(2021)Lu, Jin, Pang, Zhang, and Karniadakis]{lu2021learning}
Lu~Lu, Pengzhan Jin, Guofei Pang, Zhongqiang Zhang, and George~Em Karniadakis.
\newblock Learning nonlinear operators via {DeepONet} based on the universal approximation theorem of operators.
\newblock \emph{Nature machine intelligence}, 3\penalty0 (3):\penalty0 218--229, 2021.

\bibitem[Lu et~al.(2018)Lu, Zhong, Li, and Dong]{ode3}
Yiping Lu, Aoxiao Zhong, Quanzheng Li, and Bin Dong.
\newblock Beyond finite layer neural networks: Bridging deep architectures and numerical differential equations.
\newblock In \emph{International Conference on Machine Learning}, pages 3276--3285. PMLR, 2018.

\bibitem[Marion(2023)]{marion2023generalization}
Pierre Marion.
\newblock Generalization bounds for neural ordinary differential equations and deep residual networks.
\newblock \emph{arXiv preprint arXiv:2305.06648}, 2023.

\bibitem[Papillon et~al.(2023)Papillon, Sanborn, Hajij, and Miolane]{Papillon23}
Mathilde Papillon, Sophia Sanborn, Mustafa Hajij, and Nina Miolane.
\newblock Architectures of topological deep learning: A survey on topological neural networks.
\newblock \emph{ArXiv e-prints}, 2023.

\bibitem[Poli et~al.(2019)Poli, Massaroli, Park, Yamashita, Asama, and Park]{poli2019graph}
Michael Poli, Stefano Massaroli, Junyoung Park, Atsushi Yamashita, Hajime Asama, and Jinkyoo Park.
\newblock Graph neural ordinary differential equations.
\newblock \emph{arXiv preprint arXiv:1911.07532}, 2019.

\bibitem[Ramakrishnan et~al.(2014)Ramakrishnan, Dral, Rupp, and Von~Lilienfeld]{ramakrishnan2014quantum}
Raghunathan Ramakrishnan, Pavlo~O Dral, Matthias Rupp, and O~Anatole Von~Lilienfeld.
\newblock Quantum chemistry structures and properties of 134 kilo molecules.
\newblock \emph{Scientific data}, 1\penalty0 (1):\penalty0 1--7, 2014.

\bibitem[Rieck(2023)]{Rieck2023}
B.~Rieck.
\newblock On the expressivity of persistent homology in graph learning.
\newblock \emph{arXiv: 2302.09826}, 2023.

\bibitem[Runge(1895)]{runge1895numerische}
Carl Runge.
\newblock {\"U}ber die numerische aufl{\"o}sung von differentialgleichungen.
\newblock \emph{Mathematische Annalen}, 46\penalty0 (2):\penalty0 167--178, 1895.

\bibitem[Sander et~al.(2022)Sander, Ablin, and Peyr{\'e}]{sander2022residual}
Michael Sander, Pierre Ablin, and Gabriel Peyr{\'e}.
\newblock Do residual neural networks discretize neural ordinary differential equations?
\newblock \emph{Advances in Neural Information Processing Systems}, 35:\penalty0 36520--36532, 2022.

\bibitem[Satorras et~al.(2021)Satorras, Hoogeboom, and Welling]{satorras2021n}
V{\i}ctor~Garcia Satorras, Emiel Hoogeboom, and Max Welling.
\newblock E (n) equivariant graph neural networks.
\newblock In \emph{International conference on machine learning}, pages 9323--9332. PMLR, 2021.

\bibitem[Shi et~al.(2020)Shi, Huang, Feng, Zhong, Wang, and Sun]{shi2020masked}
Yunsheng Shi, Zhengjie Huang, Shikun Feng, Hui Zhong, Wenjin Wang, and Yu~Sun.
\newblock Masked label prediction: Unified message passing model for semi-supervised classification.
\newblock \emph{arXiv preprint arXiv:2009.03509}, 2020.

\bibitem[Southern et~al.(2024)Southern, Wayland, Bronstein, and Rieck]{southern2024curvature}
Joshua Southern, Jeremy Wayland, Michael Bronstein, and Bastian Rieck.
\newblock Curvature filtrations for graph generative model evaluation.
\newblock \emph{Advances in Neural Information Processing Systems}, 36, 2024.

\bibitem[Thomas et~al.(2018)Thomas, Smidt, Kearnes, Yang, Li, Kohlhoff, and Riley]{thomas2018tensor}
Nathaniel Thomas, Tess Smidt, Steven Kearnes, Lusann Yang, Li~Li, Kai Kohlhoff, and Patrick Riley.
\newblock Tensor field networks: Rotation-and translation-equivariant neural networks for 3d point clouds.
\newblock \emph{arXiv preprint arXiv:1802.08219}, 2018.

\bibitem[Thorpe et~al.(2022)Thorpe, Nguyen, Xia, Strohmer, Bertozzi, Osher, and Wang]{thorpe2022grand++}
Matthew Thorpe, Tan~Minh Nguyen, Heidi Xia, Thomas Strohmer, Andrea Bertozzi, Stanley Osher, and Bao Wang.
\newblock Grand++: Graph neural diffusion with a source term.
\newblock In \emph{International Conference on Learning Representation (ICLR)}, 2022.

\bibitem[Velickovi{\'c} et~al.(2017)Velickovi{\'c}, Cucurull, Casanova, Romero, Lio, and Bengio]{velivckovic2017graph}
Petar Velickovi{\'c}, Guillem Cucurull, Arantxa Casanova, Adriana Romero, Pietro Lio, and Yoshua Bengio.
\newblock Graph attention networks.
\newblock \emph{arXiv preprint arXiv:1710.10903}, 2017.

\bibitem[Verma et~al.(2022)Verma, Kaski, Heinonen, and Garg]{modflow}
Yogesh Verma, Samuel Kaski, Markus Heinonen, and Vikas Garg.
\newblock Modular flows: Differential molecular generation.
\newblock \emph{arXiv preprint arXiv:2210.06032}, 2022.

\bibitem[Verma et~al.(2023)Verma, Heinonen, and Garg]{pmlr-v202-verma23a}
Yogesh Verma, Markus Heinonen, and Vikas Garg.
\newblock {A}b{ODE}: Ab initio antibody design using conjoined {ODE}s.
\newblock In \emph{Proceedings of the 40th International Conference on Machine Learning}, pages 35037--35050. PMLR, 2023.

\bibitem[Verma et~al.(2024)Verma, Heinonen, and Garg]{verma2024climode}
Yogesh Verma, Markus Heinonen, and Vikas Garg.
\newblock Clim{ODE}: Climate and weather forecasting with physics-informed neural {ODE}s.
\newblock In \emph{The Twelfth International Conference on Learning Representations}, 2024.

\bibitem[Weinan(2017)]{ode1}
Ee~Weinan.
\newblock A proposal on machine learning via dynamical systems.
\newblock \emph{Communications in Mathematics and Statistics}, 1\penalty0 (5):\penalty0 1--11, 2017.

\bibitem[Weisfeiler and Leman(1968)]{weisfeiler1968reduction}
Boris Weisfeiler and Andrei Leman.
\newblock The reduction of a graph to canonical form and the algebra which appears therein.
\newblock \emph{nti, Series}, 2\penalty0 (9):\penalty0 12--16, 1968.

\bibitem[Xu et~al.(2019)Xu, Hu, Leskovec, and Jegelka]{gin}
K.~Xu, W.~Hu, J.~Leskovec, and S.~Jegelka.
\newblock How powerful are graph neural networks?
\newblock In \emph{International Conference on Learning Representations (ICLR)}, 2019.

\bibitem[Yildiz et~al.(2019)Yildiz, Heinonen, and Lahdesmaki]{yildiz2019ode2vae}
Cagatay Yildiz, Markus Heinonen, and Harri Lahdesmaki.
\newblock {ODE2VAE}: {D}eep generative second order {ODE}s with {B}ayesian neural networks.
\newblock \emph{NeurIPS}, 2019.

\bibitem[Zaheer et~al.(2017)Zaheer, Kottur, Ravanbakhsh, Poczos, Salakhutdinov, and Smola]{deepsets}
Manzil Zaheer, Satwik Kottur, Siamak Ravanbakhsh, Barnabas Poczos, Russ~R Salakhutdinov, and Alexander~J Smola.
\newblock Deep sets.
\newblock In \emph{Advances in Neural Information Processing Systems (NeurIPS)}, volume~30, 2017.

\end{thebibliography}
\bibliographystyle{plainnat}

\newpage
\appendix
\onecolumn

\section{Persistent homology}\label{sec:ph}

Persistent homology (PH) stands as a cornerstone in topological data analysis (TDA). At its core, PH seeks to capture multiresolution topological features (e.g., connected components, loops, voids, etc.) from data. Here, we offer a short overview of PH and direct readers to \citep{Hensel21} and \citep{ComputationalTopoBook} for an exhaustive treatment.

In the following, we consider topological spaces given by simplicial complexes. In particular, consider a simplicial complex denoted by \( K \). The $p$-chains are formal sums \( c = \sum a_i \sigma_i \), where \( a_i \in \mathbb{Z}/2\mathbb{Z} \) and \( \sigma_i \) represent $p$-dimensional simplices in \( K \). By equipping $p$-chains with addition, we obtain the group \( C_p(K) \).
Another important notion is that of boundary of a simplex. Consider a $p$-simplex \( \sigma = [v_0, ..., v_p] \in K \). The boundary of \( \sigma \) corresponds to the sum of its $(p-1)$-dimensional faces, i.e.,
\[
\partial_p \sigma = \sum_{j = 0}^{p} [v_0, ..., v_{j-1}, v_{j+1}, \ldots, v_p].
\]

Importantly, we can extend this definition to define the boundary homomorphism \( \partial_p: C_p(K) \rightarrow C_{p-1}(K) \), where \( \partial_p \sum a_i \sigma_i = \sum a_i \partial_p \sigma_i \). Then, we can define a sequence of groups, also called a chain complex, as:
\[
... C_{p+1}(K) \xrightarrow{\partial_{p+1} } C_{p}(K) \xrightarrow{\partial_{p} } C_{p-1}(K) ...
\]
where groups are connected via boundary homomorphisms. 
The $p$-th homology group comprises $p$-chains with empty boundaries (i.e., \( \partial_p\sigma = 0 \)), whereby each of these specific $p$-chains (cycles) represents a boundary of a distinct simplex in \( C_{p+1}(K) \). Hence, we define the $p$-th homology group $H_p$ as the quotient space:
\[
H_p = \mathrm{ker} \partial_p \mathrm{/} \mathrm{Im} \partial_{(p+1)}.
\]
The $p$-th Betti number of $K$, denoted by \( \beta_p \), is equal to the rank of \( H_p \). 

In persistent homology, we keep track of the evolution of Betti numbers across a sequence of chain complexes. The sequence of complexes arise from a filtration --- a nested sequence of simplicial subcomplexes \( \emptyset \subset K_{\alpha_1} \subset \ldots \subset K_{\alpha_n} = K \), indexed by timestamps $\alpha_i$ (with $\alpha_{i+1} > \alpha_i$ for all $i$). By computing the homology groups for each of these simplicial complexes, we obtain detailed topological information from $K$. In practice, this is done by associating a pair of timestamps \( (\alpha_i, \alpha_j) \) for every element of the homology groups (or topological features), indicating the filtration timestamp at which it emerged and disappeared. The persistence of a point \( (\alpha_i, \alpha_j) \) denotes the duration for which the corresponding feature persisted. We set \( \alpha_j=\infty \) if the topological feature persists until the final filtration timestamp. Formally, let $Z_p(K_{\alpha_i})=\mathrm{ker} \partial_p^{\alpha_i}$ and $B_p(K_{\alpha_i})=\mathrm{Im} \partial^{\alpha_i}_{p}$ be the standard $p$-cycle and $p$-boundary groups for the complex $K_{\alpha_i}$. Then, the $p$th persistent homology groups are
\[
H_p^{i, j} = Z_p(K_{\alpha_i}) \mathrm{/} (B_{p+1}(K_{\alpha_j}) \cap Z_p(K_{\alpha_i}))
\]
for all $1 \leq i \leq j \leq n$. Again, the $p$-th persistent Betti number \( \beta_p^{i,j} \) corresponds to the rank of \( H_p^{i,j} \). Finally, a persistence diagram comprising the persistence pairs \( (\alpha_i, \alpha_j) \) with corresponding multiplicities given by
$
\mu_p^{i,j} =(\beta_p^{i,j-1} - \beta_p^{i,j}) -  (\beta_p^{i-1,j-1} - \beta_p^{i-1,j})
$ encodes the persistent homology groups.

\section{Implementation Details}
\label{sec:implement}
Below are the implementation details.
\subsection{Graph Classification}
We followed the following hyperparameters  and training setup in \autoref{tab:hyper_error} to conduct our experiments on real-world graph classification. 

\begin{table}[!hbt]
\caption{Default hyperparameters for TopNets for Graph Classification Benchmark}
\label{tab:hyper_error}
\begin{center}
\begin{tabular}{rlc}
\toprule
Hyperparameter & Meaning & Value \\
\midrule
Solver & ODE-Solver & \texttt{adaptive-heun,euler} \\
GNN & GNN Architecture & \{GCN,GIN,MPSN\} \\
PH & Type of PH & \{VC,TOGL,RePHINE\} \\
Steps & Number of steps for ODE solver & \{20,15,10,5\} \\
Node Hidden Dim & Latent dimension of node features & 128 \\
PH embed dim & Latent dimension of PH features & 64 \\
Num Filt & Number of filtrations & 8  \\
Hiden Filtration &  Hidden dimension of filtration functions  & 16  \\
Batch Size & Size of batches  & 64\\
LR & Learning Rate & 0.001  \\
Scheduler & Learning Rate scheduler & \texttt{Cosine-Annealing-LR}  \\
Epochs & Number of epochs &  300  \\
\bottomrule
\end{tabular}
\end{center}
\end{table}

\subsection{Molecular Data QM9}
For the discrete case, we followed the data-preparation strategies, training setup, and hyperparameters as outlined by \citet{eijkelboom2023mathrm}. We enhanced each layer with an Equivariant RePHINE layer, inspired by the original RePHINE~\citep{rephine}, incorporating Euclidean distance as an invariant feature in the filtration function. The Vertex Cloud (VC) retained its absence of 3D positional information, consistent with \citep{rephine}. For the continuous case, we employed a single layer of EMPSN to parameterize the ODE dynamics, leveraging the odeint package to solve these dynamics. Additionally, an Equivariant RePHINE layer was applied per time step. Solver options included \texttt{euler} and \texttt{adaptive-heun}, with the number of time steps ranging from 5 to 20. Filtration parameters remained consistent with those described in Table \ref{tab:hyper_error}, alongside identical training hyperparameters and setup as in the original EPMSN paper. 

\subsection{CDR-H3 Antibody Design}
We followed the following hyperparameters to conduct our experiments on CDR-H3 Antibody Design.  

\begin{table}[!hbt]
\caption{Default hyperparameters for TopNets for CDR-H3 Antibody Design}
\label{tab:hyper_CDRH3}
\begin{center}
\begin{tabular}{rlc}
\toprule
Hyperparameter & Meaning & Value \\
\midrule
GNN & GNN Architecture & TransformerConv~\citep{shi2020masked} \\
PH & Type of PH & \{VC,RePHINE\} \\
Layers & Number of layers & 4 \\
Node Hidden Dim & Latent dimension of node features & [128,256,128,64] \\
PH embed dim & Latent dimension of PH features & 64 \\
Num Filt & Number of filtrations & 8  \\
Hiden Filtration &  Hidden dimension of filtration functions  & 16  \\
Batch Size & Size of batches  & 32\\
LR & Learning Rate & 0.001  \\
Scheduler & Learning Rate scheduler & \texttt{Cosine-Annealing-LR}  \\
Epochs & Number of epochs &  1000  \\
\bottomrule
\end{tabular}
\end{center}
\end{table}

\section{Deduction from TopNets}\label{sec:topnets_deduce}
In our study we restrict ourselves to $1$-dim simplicial complexes (otherwise mentioned) and here we showcases the deductions of various methods from TopNets. 

$$
\mathrm{TopAgg}^{\text{TOGL}} =\begin{cases} \mathrm{TopAgg}_{0}(\tilde{x}^\ell_\sigma, r^\ell_\sigma)= \tilde{x}_\sigma + r^\ell_\sigma, \\
\mathrm{TopAgg}_{1}(\tilde{x}^\ell_\sigma, r^\ell_\sigma)= \tilde{x}^{\ell}_\sigma\\
\mathrm{Agg}_{\ell, 0}(\{r^\ell_\sigma\}): \text{NA}\\
\mathrm{Agg}_{\ell, 1}(\{r^\ell_\sigma\})= \mathrm{DeepSet}_\ell(\{r^\ell_\sigma\}) 
\end{cases}
$$

The readout layers for TOGL concatenate the aggregated topological embeddings ($1$-dim ) with the last layer pooled $0$-dim simplex features and using it for downstream tasks such as classification.

In case of PersLay, they does not use any TNN layers over the node features, thus $\mathrm{TopAgg}_{0,1} $ are N/A, and the other aggregation is performed as, 
$$
\mathrm{TopAgg}^{\text{PersLay}} =\begin{cases}
\mathrm{Agg}_{0}(\{r_\sigma\})= \mathrm{DeepSet}_0(\{r_\sigma\}), \\
\mathrm{Agg}_{1}(\{r_\sigma\})= \mathrm{DeepSet}_1(\{r_\sigma\})
\end{cases}
$$
    
However PersLay, utilizes an additional option to use the pooled  $0$-dim simplex features via concatenating it with the aggregated topological embeddings ($0$-dim and $1$-dim) and using it for downstream tasks such as classification.

$$
\mathrm{TopAgg}^{\text{RePHINE}} =\begin{cases} \mathrm{TopAgg}_{0}(\tilde{x}^\ell_\sigma, r^\ell_\sigma)= \tilde{x}_\sigma, \\
\mathrm{TopAgg}_{1}(\tilde{x}^\ell_\sigma, r^\ell_\sigma)= \tilde{x}^{\ell}_\sigma,\\
\mathrm{Agg}_{\ell, 0}(\{r^\ell_\sigma\})= \mathrm{DeepSet}_\ell(\{r^\ell_\sigma\})\\
\mathrm{Agg}_{\ell, 1}(\{r^\ell_\sigma\})= \mathrm{DeepSet}_\ell(\{r^\ell_\sigma\}) \\
\end{cases}
$$

The readout layers for RePHINE concatenate the aggregated topological embeddings ($0$-dim and $1$-dim) with the last layer pooled $0$-dim simplex features and using it for downstream tasks such as classification.


Note that wherever we utilise $\mathrm{TopAgg}^{\mathrm{TOGL/RePHINE/PersLay}}$ as the topological aggregation method we utilise their specific readout layers as well. The \autoref{tab:deduce_ph} summarizes the deduction further from TopNets for various methods. 

\begin{table}[!hbt]
\caption{Deduction of PH-based methods from TopNets}
\label{tab:deduce_ph}
\begin{center}
\resizebox{0.8\textwidth}{!}{\begin{tabular}{rlccc}
\toprule
Module & Meaning & TOGL & PersLay & RePHINE \\
\midrule
$\mathrm{TNNLayer}$ & TNN/GNN Architecture & \{GCN,GIN\} & - & \{GCN,GIN\} \\
$\mathrm{PD}$ & Type of PH-diagrams used & VC & VC, Point transformations & RePHINE \\
$f^{\ell}$ & Filtration functions & $f_{v}$ & $f_{v}$ & $(f_{v}^{\ell},f_{e}^{\ell})$ \\
$\psi$ & Diagram combining functions &  DeepSets  &  DeepSets   & DeepSets \\
$\mathrm{TopAgg}$  & Topological Aggregation  & $\mathrm{TopAgg}^{\text{TOGL}}$ & $\mathrm{TopAgg}^{\text{PersLay}}$ & $\mathrm{TopAgg}^{\text{RePHINE}}$ \\
\bottomrule
\end{tabular}}
\end{center}
\end{table}

\section{Proofs}

\subsection{Proof of \autoref{prop:TNNs_PH}}

Let us first introduce two important notions of neighborhood for simplicial complexes: the boundary-adjacency and the upper-adjacency neighborhoods.
Let $\sigma$ be a simplex. Then, the boundary neighborhood of $\sigma$ is given by $\mathcal{B}(\sigma)=\{\tau \subset \sigma: \mathrm{dim}(\tau)=\mathrm{dim}(\sigma)-1  \}$ --- the set of $\sigma$'s faces of dimension $\mathrm{dim}(\sigma)-1$. The upper-adjacency neighborhood of $\sigma$ is $\mathcal{N}_\uparrow(\sigma)=\{\sigma' : \exists \delta \text{ such that }  \sigma \subset \delta,  \sigma' \subset \delta \text{ and } \mathrm{dim}(\delta)-1=\mathrm{dim}(\sigma')=\mathrm{dim}(\sigma)\}$ --- i.e., there exists a simplex $\delta$ that is co-face of both $\sigma$ and $\sigma'$ with dimension equal to $\mathrm{dim}(\sigma') + 1$. 

Consider simplices of a graph ($1$-dim complex). If $\sigma$ is a vertex, it has no boundary neighborhood and its upper-adjacency neighborhood are the vertices directly connected to $\sigma$. On the other hand, if $\sigma$ is an edge, it has no upper-adjacency neighborhood and its boundary one is given by the vertices that $\sigma$ is incident to.

The simplicial Weisfeiler-Leman test \citep{bodnar2021weisfeiler} resembles the original 1-WL test but takes into account the colors of the simplices of both boundary adjacency and upper adjacency in the hash (aggregating) function. Every simplex has an associated color. For a proper definition, we refer to \citet{bodnar2021weisfeiler}.

\begin{figure}[!ht]
    \centering
    \includegraphics[width=0.6\textwidth]{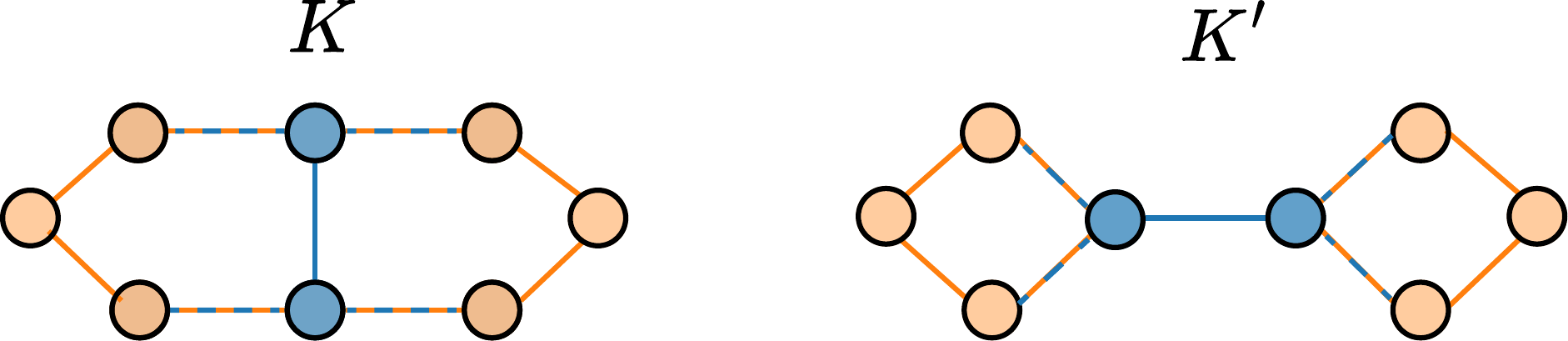}
    \caption{Two non-isomorphic simplicial complexes.}
    \label{fig:proof-prop1}
\end{figure}

To prove \autoref{prop:TNNs_PH}, it suffices to i) show a pair of clique complexes that SWL cannot distinguish, ii) and derive a color-based filtration that produces different persistence diagrams. Consider the clique complexes $K$ and $K'$ in \autoref{fig:proof-prop1}. 

We know that the multisets of colors of 0-simplices (vertices) from $K$ and $K'$ are identical at any iteration of the WL algorithm. This stems from the fact that these graphs are known to be indistiguashable by 1-WL and that the only valid neighborhood structure for vertices is the classic one (adjacent vertices) --- upper-adjacency neighborhood. In other words, for each vertex in $v \in K$ with computation tree $T_v$, there is a corresponding vertex $v' \in K'$ such that $T_v$ is isomorphic to $T_{v'}$ for any depth. 
We also note that, in SWL, the color-refinement procedure for a vertex $v$ from upper-adjacency includes the color of the edge that $v$ is incident to.
However, in our example, the color of each edge is fully defined by the history of colors of its incident vertices.
Thus, we can disregard the colors of $0$-simplices.

Similarly, if $\sigma=[u, v]$ is an edge, its only neighbors are $u$ and $v$ (boundary adjacency). If we consider edges of the same colors in $K$ and $K'$, their neighbors have isomorphic computation trees. As a result, at every iteration of the test, the colors used to update these edges are exactly the same. Therefore, SWL cannot distinguish these complexes. As noted by \citet{bodnar2021weisfeiler}, when SWL is applied to 1-simplicial complexes, i.e. graphs, it corresponds to the 1-WL test.

To prove that there exists a color-based filtration that distinguishes these graphs. We can directly leverage Theorem 2 in \citep{rephine} to show that there is a color-disconnecting set to these graphs $Q=\{\text{blue}\}$. If we remove the blue edges from $K$ and $K'$, they end up with different numbers of connected components. This concludes the proof.

\subsection{Proof of \autoref{prop:invariance_geometric_filtrations}}
Consider a geometric simplicial complex $(K, x, z)$ and geometric $i$-simplex-color filtrations induced by a function $f$. Let $R \in \mathrm{E}(n)$ be a group element that acts on the $0$-simplex positional features. Recall that geometric $i$-simplex-color filtrations leverage a function $\mathrm{Inv}(\cdot)$, which is invariant to $E(n)$ group actions. Thus, for any simplex $\tau$, we have that $\mathrm{Inv}(\{z_v\}_{v \in \tau}))=\mathrm{Inv}(R \cdot \{ z_v\}_{v \in \tau}))$. The diagrams of any dimension are fully determined by the filtrations, which in turn are obtained from the simplex rank function $o_f(\sigma)$ as 
\begin{align}
    o_f(\sigma) =\begin{cases} \underset{\tau \subset \sigma: \mathrm{dim}(\tau)=i}{\max}  f(x_\tau, \mathrm{Inv}(\{z_v\}_{v \in \tau}))&  \text{if } \mathrm{dim}(\sigma) \geq i \\
0 & \text{otherwise},
\end{cases}
\end{align}

Note that group actions only affect geometric $i$-simplex-color filtrations via the input of the $\mathrm{Inv}$ function. Thus, we can write  
\begin{align}
    R \cdot o_f(\sigma) =\begin{cases} \underset{\tau \subset \sigma: \mathrm{dim}(\tau)=i}{\max}  f(x_\tau, \mathrm{Inv}(R\cdot\{z_v\}_{v \in \tau}))&  \text{if } \mathrm{dim}(\sigma) \geq i \\
0 & \text{otherwise},
\end{cases}
\end{align}

Recalling that invariant features remain intact via the transformation, this would imply $o_f(\sigma) = R \cdot o_f(\sigma)$, which would lead to identical filtrations and, consequently, the same persistence diagrams (of any dimension) and topological embeddings. This holds for any $i \geq 0$.

\section{Approximation Error Bounds}
\label{sec:bounds}
Below are the bounds for the TOGL and RePHINE cases. Note that we assume a fixed simplicial complex for deriving the bound.

\subsection{TOGL}
\label{subsec:togl_bound}

\subsubsection{Continuous Counterpart}
The dynamics of the TOGL-GNN for a node $v$ can be described as,
\begin{align}
x_{v}^\ell =  \mathrm{TNNLayer}_{\ell}(x_{v}^{\ell-1},K) +   \psi(\mathrm{PD}(\sigma; f_\theta, x_{v}^{\ell-1}, K))
\end{align}
For clarity of exposition, let $\mathrm{TNNLayer}_{\ell}(x^{\ell-1},K) = x_{v}^{\ell-1} + m^{\ell}_v$, where $m^{\ell}_v$ is the aggregated message as described in Section~\ref{sec:cont_equiv}. The continuous depth counterpart can be written as a graph ODE, parametrized by the following differential equation,
\begin{align}
    \dot{x}_{v}^{t} = \psi(\mathrm{PD}(\sigma; f_\theta, x^{t}, K)) + m^t_v
\end{align}

\subsubsection{Error Bound}
We consider N-layered TOGL GNN and assume an Euler discretization scheme for the ODE system consisting of $N$ steps to be consistent. We define $s_{\ell} = \nicefrac{\ell}{N} = \ell h$, where $h=\nicefrac{1}{N}$ is the step size and, $s_{\ell}$ represents a time at $\ell^{th}$ step.  We utilize the Taylor expansion as,
\begin{align}
    x_{v}^{s_{\ell+h}} = x_{v}^{s_{\ell}} + h\dot{x}_{v}^{s_{\ell}} + R_{1}(h)
\end{align}
We consider a simple modification of the discrete TOGL GNN network for $N$-depth by letting the  mapping explicitly depend on the depth of the network as,
\begin{align}
    x_{v}^\ell  =x_{v}^{\ell-1} + \frac{1}{N} \left(\psi(\mathrm{PD}(\sigma; f_\theta, x^{\ell-1}, K)) + m^{\ell-1}_v\right)
\end{align}
We consider the error $e_v(\ell) = x_{v}^{s_{\ell}} -  x_{v}^\ell$, where $x_v^\ell$ is the node $v$ embeddings after $l$ TOGL-GNN layers,
\begin{align}
    e_v(\ell+1) - e_v(\ell) &=  x_{v}^{s_{\ell+1}} -  x_{v}^{s_{\ell}} + x_{v}^\ell - x_{v}^{\ell+1}\\
    &= h\dot{x}_{v}^{s_{\ell}} + R_{1}(h) -\frac{1}{N} \left(\psi(\mathrm{PD}(\sigma; f_\theta, x^{\ell}, K)) + m_{v}^{\ell}\right) \\
    &= R_{1}(h) + h\left(m_{v}^{s_{\ell}}  - m_{v}^{\ell}\right) \\&
    ~~~~~~+ h\left( \psi(\mathrm{PD}(\sigma; f_\theta, x^{s_{\ell}}, K)) -  \psi(\mathrm{PD}(\sigma; f_\theta, x^{\ell}, K))\right)
\end{align}
We assume $m_{v}$,$\psi$ to be $L_{m},L_{\beta}$-Lipschitz ($L_{\beta} = L_{\psi}L_{\theta}$, due to the composition of $\psi$ and $f_{\theta}$), giving us, (note that the parametrization of $m_{v}^{\ell}$ and $m^{s_{\ell}}_v$ is the same, and only differs in inputs.)
\begin{align}
    \|e_v(\ell+1) - e_v(\ell)\| &\leq R_{1}(h) + hL_{m}\|e_v(\ell)\| + hL_{\beta}\|e_v(\ell)\|\\
    \|e_v(\ell+1)\| &\leq R_{1}(h) + \left(1 + \frac{L_{m} + L_{\beta}}{N}\right)\|e_v(\ell)\| 
\end{align}
Using the discrete Gronwall lemma~\citep{sander2022residual,demailly2006analyse}, we get the following relation, where $e_v(0) = 0$,
\begin{align}
    \|e_v(\ell)\| &\leq 0 + R_{1}(h) \sum_{0 \leq j \leq n-1} \exp(\frac{L_{m} + L_{\beta}}{N}(N-1-j)) \\
    &\leq R_{1}(h) \frac{\exp(\frac{L_{m} + L_{\psi}}{N}N) -1}{\exp(\frac{L_{m} + L_{\beta}}{N}) -1}
\end{align}
But, $\exp(\frac{L_{m} + L_{\beta}}{N}) -1 \geq \frac{L_{m} + L_{\beta}}{N}$, using that we get,
\begin{align}
    \|e_v(\ell)\| &\leq R_{1}(h)\frac{N(\exp(L_{m} + L_{\beta}) -1)}{L_{m} + L_{\beta}}
\end{align}

\subsection{RePHINE}
\label{subsec:rephine_bound}
\subsubsection{Continuous Counterpart}
The dynamics of RePHINE-GNN for node $v$ can be expressed as,
\begin{align}
x_{v}^\ell &=  \mathrm{TNNLayer}_{\ell}(x_{v}^{\ell-1},K)\\
r^{\ell} &= \psi^{\ell}(\mathrm{PD}(\sigma; f^{\ell}_\theta, x^{\ell-1}, K))
\end{align}
Let $\mathrm{TNNLayer}_{\ell}(x_{v}^{\ell-1},K) = x_{v}^{\ell-1} + m_{v}^{\ell}$, where $m^{\ell}_v$ is the aggregate as described in Section~\ref{sec:cont_equiv} and  $x^{\ell} = \{ x_{u}^{\ell}\}_u$. Moreover, collecting all node updates, the recursive update can be expressed as $x^{\ell} = x^{\ell-1} + m^{\ell}$, where $m^{\ell}$ are the message updates for the all node embeddings. RePHINE parameterizes each layer filtration function $f^{\ell}_\theta$ and DeepSet function $\psi^{\ell}$ distinctively. The continuous depth counterpart can be written as a coupled latent graph ODE, parametrized by the following set of differential equations as,
\begin{align}
    \dot{x}_{v}^{t} &= m_{\sigma}^{t} \\
    r^{t} &= \psi^{t}(\mathrm{PD}(\sigma; f^{t}_\theta, x^{t}, K))
\end{align}

\subsubsection{Error Bound}
We consider N layered RePHINE GNN, and assume an Euler discretization scheme for the ODE system consisting of $N$ steps to be consistent. We define $s_{\ell} = \nicefrac{\ell}{N}= \ell h$, where $h=\frac{1}{N}$ is the step size and, $s_{\ell}$ represents a time at $\ell^{th}$ step. We derive the error bounds both for the node features and topological embeddings as follows.

\paragraph{Node Embeddings}

We utilize the Taylor expansion, as,
\begin{align}
    x_{v}^{s_{\ell+h}} = x_{v}^{s_{\ell}} + h\dot{x}_{v}^{s_{\ell}} + R_{1}(h)
\end{align}
We consider a simple modification of the discrete RePHINE GNN network for $N$-depth by letting the  mapping explicitly depend on the depth of the network as,
\begin{align}
    x_{v}^\ell  =x_{v}^{\ell-1} + \frac{1}{N} m^{\ell}_v
\end{align}
We consider the node-embedding error, $e_{v}(\ell) = x_{v}^{s_{\ell}} - x_{v}^\ell$, 
\begin{align}
    e_{v}(\ell+1) - e_{v}(\ell) &= x_{v}^{s_{\ell+1}} - x_{v}^{s_{\ell}} + x_{v}^\ell - x_{v}^{\ell+1}\\
    &= h\dot{x}_{v}^{s_{\ell}} + R_{1}(h) -  \frac{1}{N}m^{\ell}_v \\
    &=  R_{1}(h) + h\left(m^{s_{\ell}}_v - m^{\ell}_v \right)
\end{align}
Assuming $m_{\sigma}$ to be $L_{m}$-Lipschitz, gives us (note that the parametrization of $m_{\sigma}^{s_{\ell}}$ and $m^{\ell}_\sigma$ is the same, and only differs in inputs.)
\begin{align}
    \|e_{v}(\ell+1) - e_{v}(\ell)\| &\leq R_{1}(h) + hL_{m}||e_{v}(\ell)|| \\
    \|e_{v}(\ell+1)\| &\leq R_{1}(h) + \left(1 + \frac{L_{m}}{N}\right)||e_{v}(\ell)|| 
\end{align}
Using the discrete Gronwall lemma, we get the following relation, where $\e_{x}(0) = 0$,
\begin{align}
    \|e_{v}(\ell)\| &\leq 0 + R_{1}(h) \sum_{0 \leq j \leq n-1} \exp(\frac{L_{m}}{N}(N-1-j)) \\
    &\leq R_{1}(h) \frac{\exp^(\frac{L_{m}}{N} N) -1}{\exp(\frac{L_{m}}{N}) -1}
\end{align}
But, $\exp(\frac{L_{m}}{N}) -1 \geq \frac{L_{m}}{N}$, using that we get,
\begin{align}
    \|e_{v}(\ell)\| &\leq R_{1}(h)\frac{N(\exp(L_{m}) -1)}{L_{m}}
\end{align}

\paragraph{Topological Embeddings} We consider the bound on topological-embedding in this section. Let $r^{\ell}$ is the topological-embedding after $l$ RePHINE-GNN layers, the error bound can be computed as,
\begin{align}
    e_{r}(\ell) &= r^{s_{\ell}}(x^{s_{\ell}},K) - r^{\ell}(x^{\ell},K)\\
    &  = r^{s_{\ell}}( x^{s_{\ell-1}}) + h\frac{d r^{s_{\ell}}(x^{s_{\ell-1}}  )}{d x^{s_{\ell-1}}  }m^{s_{\ell-1}}  + R_{1}(h) - r^{\ell}(x^{\ell-1} + \frac{1}{N}m^{\ell-1})
\end{align}
Now, using the Taylor expansion to expand the second term, we can write ($h=1/N$)
\begin{align}
    r^{\ell}(x^{\ell-1} + \frac{1}{N}m^{\ell-1}) =  r^{\ell}(x^{\ell-1}) + h\frac{d r^{\ell}(x^{\ell-1})}{d x^{\ell-1}}m^{\ell-1} + R_{1}(m^{\ell-1}) 
\end{align}

Putting into the original equation, we get,
\begin{align}
    e_{r}(\ell) &= \underbrace{r^{s_{\ell}}(x^{s_{\ell-1}}) - r^{\ell}(x^{\ell-1})}_\text{First Term} + \underbrace{h\frac{d r^{s_{\ell}}(x^{s_{\ell-1}})}{d x^{s_{\ell-1}}  }m^{s_{\ell-1}} - h\frac{d r^{\ell}(x^{\ell-1})}{d x^{\ell-1}}m^{\ell-1}}_\text{Second Term} \\
    &~~~~~~~~~~+ R_{1}(h) - R_{1}(m^{\ell-1})
\end{align}
We simplify each term as follows,

\underline{First Term}: 
The first term denotes the difference between the topological embeddings, and we assume that $\psi^{s_{\ell}} \equiv \psi^{\ell}$, as both functions are evaluated at the $\ell$ layer (step), and let it be $L_{\beta}^{\ell}$-Lipschitz ($L_{\beta}^{\ell} = L_{\psi}^{\ell}L_{\theta}^{\ell}$, due to the composition of $\psi^{\ell}$ and $f_{\theta}^{\ell}$) at the time-step, giving us
\begin{align}
    \| r^{s_{\ell}}(x^{s_{\ell-1}}) - r^{\ell}(x^{\ell-1})\| &=  \|\psi^{s_{\ell}}(\mathrm{PD}(\sigma; f^{s_{\ell}}_\theta, x^{s_{\ell}-1}, K)) - \psi^{\ell}(\mathrm{PD}(\sigma; f^{\ell}_\theta, x^{\ell-1}, K)|| \\
    & \leq L^{\ell}_{\beta} \|e_{v}(\ell-1)\|
\end{align}

\underline{Second Term}: We simplify the second term as follows, by adding and subtracting a term as,
\begin{align}
    =& h\frac{d r^{s_{\ell}}(x^{s_{\ell-1}})}{d x^{s_{\ell-1}}  }m^{s_{\ell-1}} - h\frac{d r^{s_{\ell}}(x^{s_{\ell-1}}  )}{d x^{s_{\ell-1}} }m^{\ell-1} + h\frac{d r^{s_{\ell}}(x^{s_{\ell-1}}  )}{d x^{s_{\ell-1}}  }m^{\ell-1} - h\frac{d r^{\ell}(x^{\ell-1})}{d x^{\ell-1}}m^{\ell-1} \\
    =&h\left( \frac{d r^{s_{\ell}}(x^{s_{\ell-1}}  )}{d x^{s_{\ell-1}}  } \left(m^{s_{\ell-1}} - m^{\ell-1}\right)  + m^{\ell-1}\left(  \frac{d r^{s_{\ell}}(x^{s_{\ell-1}}  )}{d x^{s_{\ell-1}}  } - \frac{d r^{\ell}(x^{\ell-1})}{d x^{\ell-1}}    \right) \right)
\end{align}
where the parts of the second term can be simplified as,
\begin{align}
    \frac{d r^{s_{\ell}}(x^{s_{\ell-1}})}{d x^{s_{\ell-1}} } & = \frac{|r^{s_{\ell}}(x^{s_{\ell-1+h}}  ) - r^{s_{\ell}}(x^{s_{\ell-1}})|}{ |x^{s_{\ell-1+h}} - x^{s_{\ell-1}}|} \leq L_{\beta}^{\ell}
\end{align}
Similarly, the other term,
\begin{align}
    \frac{d r^{\ell}(x^{\ell-1})}{d x^{\ell-1}}  = \frac{|r^{\ell}(x^{\ell-1+h}) - r^{\ell}(x^{\ell-1})|}{ |x^{\ell-1+h} - x^{\ell-1}|} \leq L_{\beta}^{\ell}
\end{align}
leading to $\leq (L_{\beta}^{\ell} - L_{\beta}^{\ell}) = 0$.
So, the equation will become,
\begin{align}
    h \frac{d r^{s_{\ell}}(x^{s_{\ell-1}}  )}{d x^{s_{\ell-1}}} \left(m^{s_{\ell-1}} - m^{\ell-1}\right) \leq&  h \frac{d r^{s_{\ell}}(x^{s_{\ell-1}}  )}{d x^{s_{\ell-1}} } L_{m}\|e_{v}(\ell-1)\| \\&\leq\frac{L_{\beta}^{\ell}L_{m}}{N} \|e_{v}(\ell-1)\| 
\end{align}

Collecting all the terms, it will account for,
\begin{align}
    \|e_{r}(\ell)\| \leq L^{\ell}_{\beta} \|e_{v}(\ell-1)\|+ \frac{L_{\beta}^{\ell}L_{m}}{N} \|e_{v}(\ell-1)\| + R_{1}(h) - R_{1}(m^{\ell-1})
\end{align}

\end{document}